\documentclass[runningheads]{llncs}
\usepackage{graphicx}

\usepackage{tikz}
\usepackage{comment}
\usepackage{amsmath,amssymb} %
\usepackage{color}
\usepackage[pagebackref=true,breaklinks=true,colorlinks,citecolor=cyan,linkcolor=red,bookmarks=false]{hyperref}
\usepackage{times}
\usepackage[misc]{ifsym}

\usepackage{epsfig}
\usepackage{booktabs}
\usepackage{graphbox} %
\usepackage{capt-of} %
\usepackage{mathtools} %
\usepackage{multirow}
\usepackage{placeins} %
\usepackage{adjustbox}
\usepackage{nicefrac}
\usepackage{overpic}
\usepackage{makecell}

\usepackage[accsupp]{axessibility}  
\usepackage{subfig}

\newcommand{\secref}[1]{Sec.~\ref{#1}}
\newcommand{\tabref}[1]{Table.~\ref{#1}}
\newcommand{\figref}[1]{Fig.~\ref{#1}}
\renewcommand{\eqref}[1]{Equ.~\ref{#1}}

\newcommand{\RR}{\mathbb{R}}
\newcommand{\codebook}{\mathcal{Z}}

\newcommand{\decoder}{G}
\newcommand{\encoder}{E}
\newcommand{\quantize}{\mathbf{q}}
\newcommand{\quantizedcode}{z_{\mathbf{q}}}

\newcommand{\myPara}[1]{\noindent\textbf{#1}}

\newcommand{\redbf}[1]{{\textbf{\color{red}{#1}}}} %
\newcommand{\blueud}[1]{{\underline{\color{blue}{#1}}}} %
\newcommand{\etal}{\textit{et al}}

\graphicspath{ {./img/}, {./suppimg/} }

\begin{document}

\pagestyle{headings}
\mainmatter
\def\ECCVSubNumber{1617}  %

\title{VQFR: Blind Face Restoration with Vector-Quantized Dictionary and Parallel Decoder} %

\titlerunning{VQFR: Vector-Quantized Face Restoration}

\author{Yuchao Gu$^{*}$\inst{1,2} \and
 Xintao Wang$^{\dagger}$\inst{2} \and 
 Liangbin Xie\inst{2,5} \and 
 Chao Dong\inst{4,5} \and 
 \\ Gen Li\inst{3} \and 
 Ying Shan\inst{2} \and 
 Ming-Ming Cheng \Letter\inst{1} \orcidID{0000-0001-5550-8758} 
 }
\authorrunning{Gu et al.}
\institute{ $^1$TMCC, CS, Nankai University \qquad $^2$ARC Lab, Tencent PCG\\ $^3$Platform Technologies, Tencent Online Video \qquad $^4$Shanghai AI Laboratory \\
 $^5$Shenzhen Institute of Advanced Technology, Chinese Academy of Sciences\\
 \url{https://github.com/TencentARC/VQFR/}
 }

\maketitle

\let\thefootnote\relax\footnotetext{$^\dagger$ Project lead. \quad
\Letter \; Corresponding author.\\
$^{*}$ Yuchao Gu is an intern in ARC Lab, Tencent PCG.}

\begin{abstract}
Although generative facial prior and geometric prior have recently demonstrated high-quality results for blind face restoration, producing fine-grained facial details faithful to inputs remains a challenging problem. 
Motivated by the classical dictionary-based methods and the recent vector quantization (VQ) technique, we propose a VQ-based face restoration method -- VQFR. VQFR takes advantage of high-quality low-level feature banks extracted from high-quality faces and can thus help recover realistic facial details. However, the simple application of the VQ
codebook cannot achieve good results with faithful details and identity preservation. Therefore, we further introduce two special network designs. 1). We first investigate the compression patch size in the VQ codebook and find that the VQ codebook designed with a proper compression patch size is crucial to balance the quality and fidelity. 2). To further fuse low-level features from inputs while not “contaminating” the realistic details generated from the VQ codebook, we proposed a parallel decoder consisting of a texture decoder and a main decoder. Those two decoders then interact with a texture warping module with deformable convolution. Equipped with the VQ codebook as a facial detail dictionary and the parallel decoder design, the proposed VQFR can largely enhance the restored quality of facial details while keeping the fidelity to previous methods.
\end{abstract}

\section{Introduction}

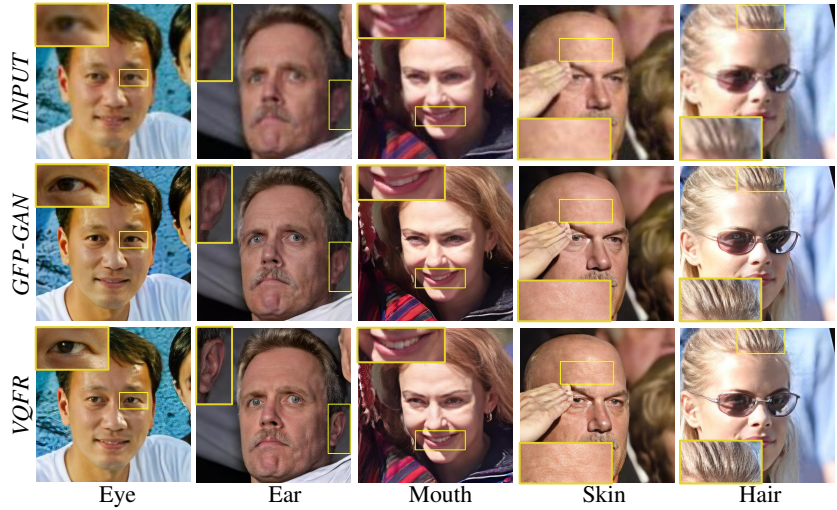
\begin{figure*}[!tb]
	\centering
	\begin{overpic}[width=0.9\textwidth]{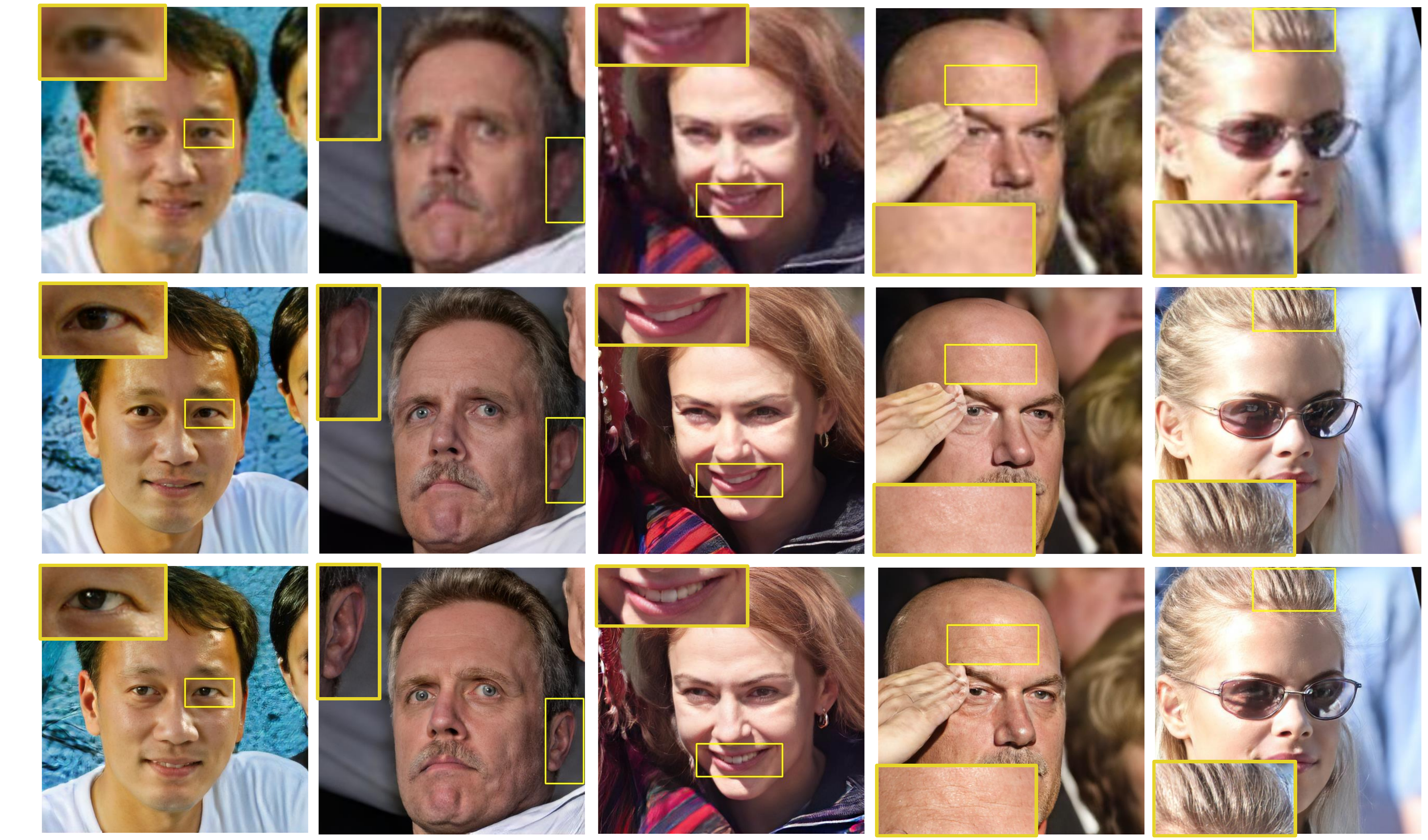}
		\put(0,44){\rotatebox{90}{\textit{INPUT}}}
		\put(0,23){\rotatebox{90}{\textit{GFP-GAN}}}
		\put(0,6){\rotatebox{90}{\textit{VQFR}}}
		\put(10.5,-2){Eye}
		\put(31,-2){Ear}
		\put(48,-2){Mouth}
		\put(69,-2){Skin}
		\put(88,-2){Hair}
	\end{overpic}
	\caption{Comparisons of restoration quality between GFP-GAN~\cite{wang2021towards} and VQFR.
		Our VQFR can restore high-quality facial details on various facial regions and keep the fidelity as well, while GFP-GAN lacks realistic fine details. (\textbf{Zoom in for best view})}
	\label{fig:facialdetails}
\end{figure*}

Blind face restoration aims at recovering low-quality (LQ) faces with unknown degradations, such as noise \cite{zhang2017beyond}, blur \cite{kupyn2018deblurgan,shen2018deep}, down-sampling~\cite{dong2014learning,lim2017enhanced}, \textit{etc}.
This task becomes more challenging in real-world scenarios, where there are more complicated degradations, diverse face poses and expressions.
Previous works typically exploit face-specific priors, including geometric priors~\cite{chen2018fsrnet,yu2018face,chen2021progressive}, generative priors~\cite{wang2021towards,yang2021gan} and reference priors~\cite{li2020blind,li2018learning,dogan2019exemplar}.
Specifically, geometric priors usually consist of facial landmarks~\cite{chen2018fsrnet}, face parsing maps~\cite{chen2018fsrnet,chen2021progressive} and facial component heatmaps~\cite{yu2018face}.
They could provide global guidance for restoring accurate face shapes, but do not help generate realistic details.
Besides, geometric priors are estimated from degraded images and thus become inaccurate for inputs with severe degradations.
These properties motivate researchers to find better priors.

Recent works~\cite{wang2021towards,yang2021gan} begin to investigate generative priors in face restoration and achieve superior performance.
They usually leverage the powerful generation ability of a pre-trained face generative adversarial network (\textit{e.g.}, StyleGAN~\cite{karras2019style,karras2020analyzing}) to generate realistic textures.
These methods typically project the degraded images back into the GAN latent space, and then decode high-quality (HQ) faces with the pre-trained generator.
Although GAN-prior-based methods achieve decent overall restoration quality at first glance, they still fail to produce fine-grained facial details, especially the fine hair and delicate facial components (see examples in \figref{fig:facialdetails}).
This can be partially attributed to the imperfect latent space of the well-trained GAN model.
Reference-based methods
explore the high-quality guided faces~\cite{dogan2019exemplar,li2018learning} or facial component dictionary~\cite{li2020blind} to solve face restoration problems.
DFDNet~\cite{li2020blind} is a representative method, which does not need to access the faces of the same identity. It explicitly establishes a high-quality ``texture bank" for several facial components and then replaces degraded facial components with the nearest HQ facial components in the dictionary.
Such a discrete replacement operation directly bridges the gap between the low-quality facial components and high-quality ones, thus having the potential to provide decent facial details.
However, the facial component dictionary in DFDNet still has two weaknesses.
1) It offline generates the facial component dictionary with a pre-trained VGGFace~\cite{cao2018vggface2} network, which is optimized for the recognition task but is sub-optimal for restoration. 2) It only focuses on several facial components (\textit{i.e.}, eyes, nose, and mouth), but does not include other important areas, such as hair and skin.

The limitations of the facial component dictionary motivate us to explore Vector Quantized (VQ) codebook, a dictionary constructed for all facial areas.
The proposed face restoration method -- VQFR, takes advantage of both dictionary-based methods and GAN training, yet does not require any geometric or GAN prior. Compared to the facial component dictionary~\cite{li2020blind}, the VQ codebook could provide a more comprehensive low-level feature bank that is not restricted to limited facial components. It is also learned in an end-to-end manner by the face reconstruction task. Besides, the mechanism of vector quantization makes it more robust for diverse degradations. Nevertheless, it is not easy to achieve good results simply by applying the VQ codebook. We further introduce two special network designs, which allow VQFR to surpass previous methods in both detail generation and identity preserving.

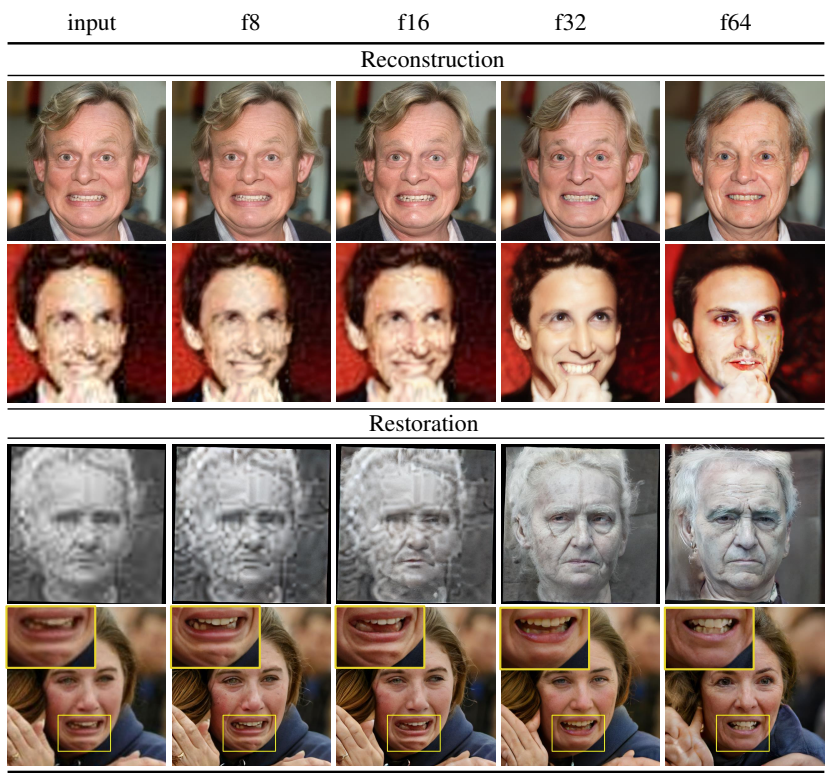
\begin{figure*}[!tb]
	\centering
	\begin{overpic}[width=0.9\textwidth]{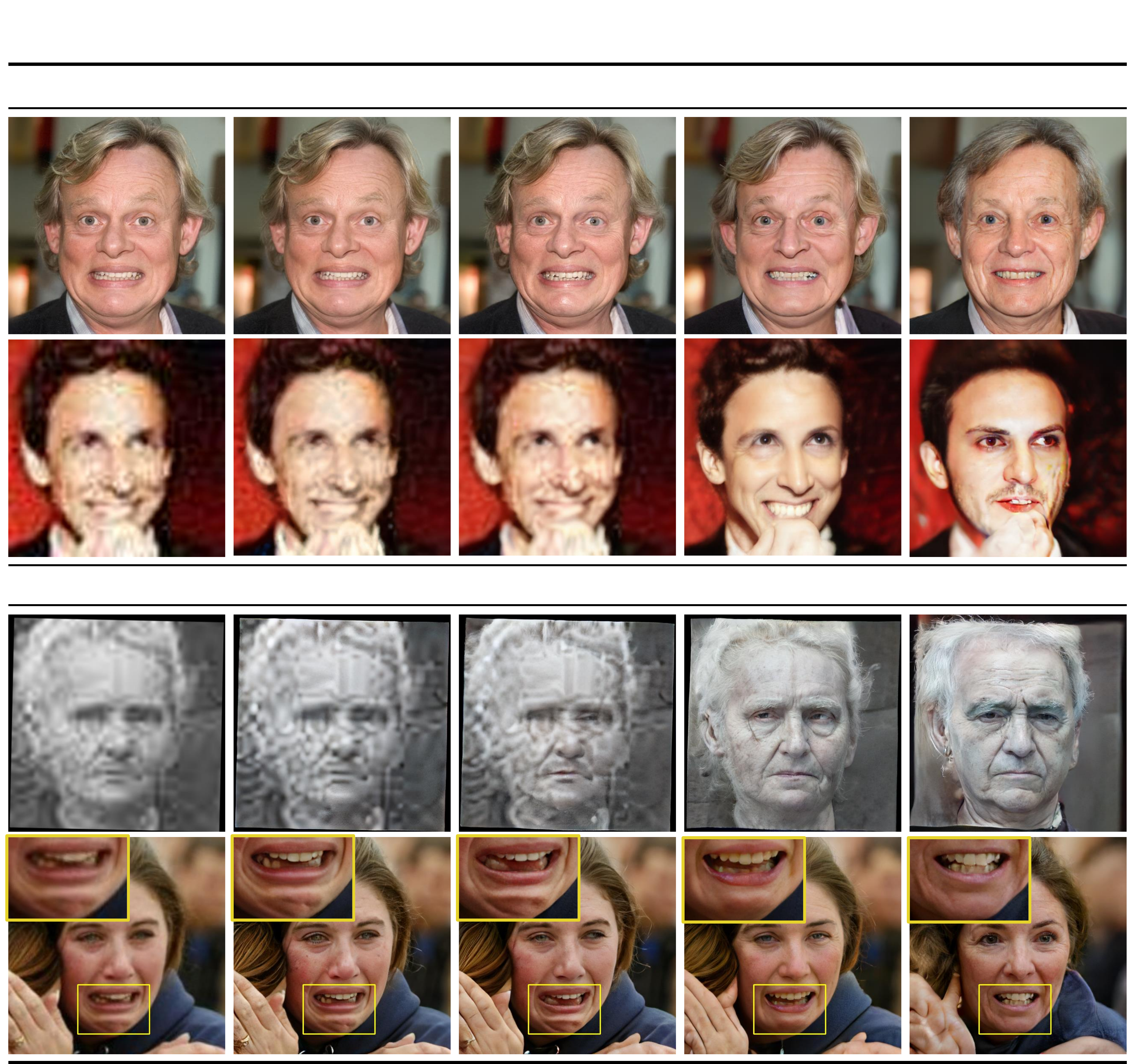}
	\put(43.5,85.5){Reconstruction}
	\put(44.5,41.4){Restoration}
		\put(8,90){input}
		\put(29,90){f8}
		\put(48,90){f16}
		\put(67,90){f32}
		\put(87,90){f64}
	\end{overpic}
	\caption{Reconstruction and restoration results based on codebook with compression patch size $f=\{8,16,32,64\}$. In the reconstruction experiment, we analyze the HQ reconstruction (row 1) and LQ reconstruction (row 2) based on pretrained HQ codebook. In the restoration experiment, we visualized the restoration result of faces of large degradation (row 3) and small degradation (row 4), respectively.   }
	\label{fig:motivation}
\end{figure*}

First, to generate realistic details, we find that it is crucial to select a proper compression patch size $f$, which indicates ``how large a patch is represented" by an atom of the codebook.  As shown in \figref{fig:motivation}, a larger $f$ could lead to better visual quality but worse fidelity. After a comprehensive investigation, we suggest using $f=32$ for the input image size $512\times 512$.
However, such a selection is only a trade-off between quality and fidelity. The expression and identity could also change a lot even with a proper compression patch size (see \figref{fig:motivation}). A straightforward solution is to fuse low-level features from input into different decoder layers, just like in GFP-GAN~\cite{wang2021towards}. Although the input features could bring more fidelity information, they will also ``contaminate" the realistic details generated from the VQ codebook. This problem leads to our second network design -- a parallel decoder.  Specifically, the parallel decoder structure consists of a texture decoder and a main decoder. The texture decoder only receives information from the latent representations from the VQ codebook, while the main decoder warps the features from the texture decoder to match the information from degraded input.
In order to eliminate the loss of high-quality details and better match the degraded faces, we further adopt a texture warping module with deformable convolution~\cite{zhu2019deformable} in the main decoder.
Equipped with the VQ codebook as a facial dictionary and the parallel decoder design, we can achieve more high-quality facial details while reserving the fidelity for face restoration.

Our contributions are summarized as follows:
\begin{enumerate}
    \item We propose the VQ dictionary of HQ facial details for face restoration. Our analysis of the VQ codebook shows the potential and limitations of the VQ codebook, together with the importance of the compression patch sizes in face restoration.
    \item A parallel decoder is proposed to gradually fuse input features and texture features from VQ codebooks, which keeps the fidelity without sacrificing HQ facial details.
    \item Extensive experiments with quantitative and qualitative comparisons show  VQFR largely surpasses previous works in restoration quality while keeping high fidelity.
\end{enumerate}
\section{Related Work}

\noindent\textbf{Blind Face Restoration}

\noindent Early works explore different facial priors in face restoration. Those priors can be categorized into three types:
geometric priors~\cite{chen2018fsrnet,yu2018face,chen2021progressive}, generative priors~\cite{wang2021towards,yang2021gan} and reference priors~\cite{li2020blind,li2018learning,dogan2019exemplar}.
The geometric priors include facial landmark~\cite{chen2018fsrnet}, face parsing maps~\cite{chen2018fsrnet,chen2021progressive} and facial component heatmaps~\cite{yu2018face}.
Those priors are estimated from degraded images and thus become inaccurate for inputs with severe degradations.
Besides, the geometric structures cannot provide sufficient information to recover facial details. In this work, we do not explicitly integrate geometric priors, but we use landmark distance to estimate restoration fidelity.

Recent works investigate the generative priors to provide facial details and achieve decent performance.
In the early arts, GAN inversion methods~\cite{menon2020pulse,gu2020image} aim to find the closest latent vector in the GAN space given an input image.
Within this category, PULSE~\cite{menon2020pulse} iteratively optimizes the latent code of a pre-trained StyleGAN~\cite{karras2019style}. mGANprior~\cite{gu2020image} simultaneously optimize several codes to promote its reconstruction.
Recent works GFP-GAN~\cite{wang2021towards} and GPEN~\cite{yang2021gan} extract fidelity information from inputs and then leverage the pre-trained GAN as a decoder, which achieves a good balance between visual quality and fidelity.
Those methods still fail to produce fine-grained facial details.
We conjecture that StyleGAN constructs a continuous latent space, and thus GAN-prior methods easily project degraded faces into a suboptimal latent code.

Reference priors~\cite{li2020blind,li2018learning,dogan2019exemplar} typically rely on reference
images of the same identity. To address this issue, DFDNet~\cite{li2020blind} constructs an offline facial component dictionary~\cite{li2020blind} with VGGFace~\cite{cao2018vggface2} for face recognition.
It then conducts discrete replacement, \textit{i.e.}, replacing the degraded facial components with high-quality ones in the dictionary by the nearest search. With the facial component dictionary, DFDNet~\cite{li2020blind} restores better facial components.
However, it still has two limitations: 1) The dictionary is offline generated with a recognition model, which is sub-optimal for face restoration.
2) It only builds on several facial components (\textit{i.e.}, eyes, nose and mouth), leaving other facial regions like skin and hair untouched. In this work, we explore the VQ codebook as a facial dictionary, which can be end-to-end trained by a reconstruction objective and provide realistic facial details.
Recent RestoreFormer~\cite{wang2022restoreformer} also exploits the VQ codebook, but their work mainly discusses the diverse cross-attention mechanism for LQ latent and HQ code interaction. 
It does not explore deeply in VQ codebook while we show the dilemma between fidelity and realness when using the VQ codebook of different scales. 
Moreover, restoreformer does not exploit the input feature of larger resolution and thus results in limited restoration fidelity. Instead, we explore the parallel decoder to achieve realness and fidelity simultaneously.

\noindent\textbf{Vector-Quantized Codebook}

\noindent The vector-quantized codebook is first introduced in VQ-VAE~\cite{van2017neural}. With this codebook, the encoder network outputs are discrete rather than continuous, and the prior encapsulated in the codebook is learned rather than static.
The following works propose different improvements to codebook learning. VQVAE2~\cite{razavi2019generating} introduces a multi-scale codebook for better image generation.
VQGAN~\cite{esser2021taming} trains the codebook with the adversarial objective and thus the codebook can achieve high perceptual quality.
To improve the codebook usage, some works~\cite{lancucki2020robust,yu2021vector} explore training techniques like L2-normalization or periodically re-initialization.
Such a VQ codebook is a patch tokenizer and can be adopted in multiple tasks, like image generation~\cite{chang2022maskgit,yu2021vector,esser2021taming}, multi-modal generation~\cite{wu2021n} and large-scale pretraining~\cite{dong2021peco,bao2021beit}.
Different from previous works that use the VQ codebook to get token features, we explore the potential of the VQ codebook as an HQ facial details dictionary.

\begin{figure*}[!tb]
	\centering
	\includegraphics[width=0.9\textwidth]{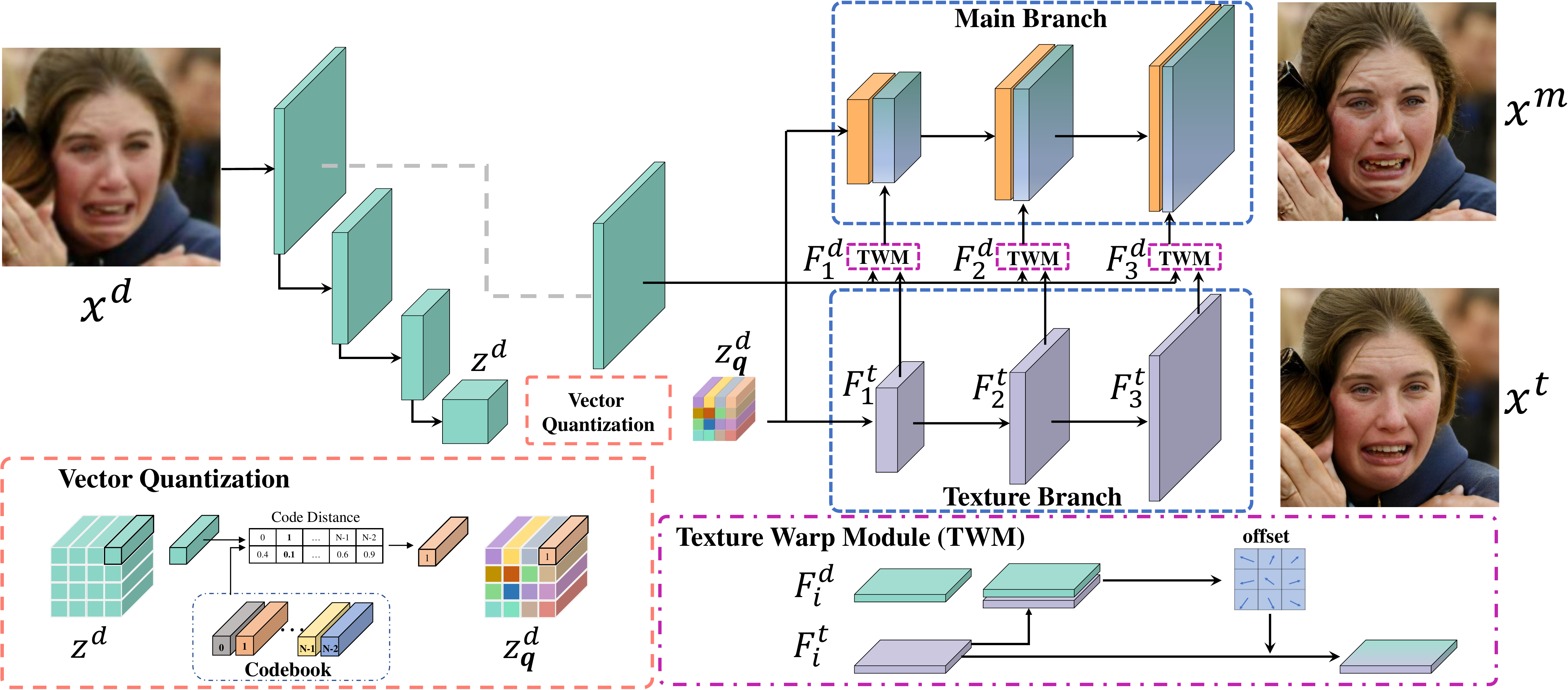}
	\caption{\textbf{Overview of VQFR framework}. It consists of an encoder to map degraded face into latent and a parallel decoder to exploit the HQ code and input feature. The encoder and decoder are bridged by vector quantization model and a pretrained HQ codebook to replace the encoded latent to HQ code.}
	\label{fig:pipeline}
\end{figure*}

\section{Methodology}

We describe the proposed VQFR framework in this section. Our goal is to restore high-quality faces with realistic facial details while reserving the fidelity of degraded faces. 
To achieve this goal, VQFR explores two key ingredients: Vector-Quantized (VQ) dictionary and parallel decoder. The overview of VQFR framework is illustrated in \figref{fig:pipeline}. 
VQFR contains an encoder, a parallel decoder and a pretrained HQ codebook. 
We first learn a VQ codebook from only HQ faces with an encoder-decoder structure by the vector-quantization technique\cite{van2017neural,esser2021taming}. Then, a degraded face is encoded to a compact latent representation with a downsampling factor $f$. At each spatial location of this latent representation, we replace the latent vector with its nearest code in the HQ codebook.
After that, the substituted latent representation is then decoded back to the image space, \textit{i.e.}, restored image with high-quality facial details.

The remaining parts are arranged as follows: 
In \secref{sec:vqcodebook}, we first analyze the potential and limitation of vector-quantization technique as a texture dictionary in face restoration task. Then we introduce the parallel decoder to promote the fidelity while maintaining the rich high-quality facial details in \secref{sec:parallel}. In \secref{sec:objective}, we describe the overall training objective of VQFR framework.

\subsection{Vector-Quantized Codebook}
\subsubsection{Preliminary.}
\label{sec:vqcodebook}
The Vector-Quantized (VQ) codebook is first introduced in VQVAE~\cite{van2017neural}, which aims to learn discrete priors to encode images. 
The following work VQGAN~\cite{esser2021taming} proposes a perceptual codebook by further using perceptual loss~\cite{johnson2016perceptual} and adversarial training objectives~\cite{isola2017image}.
We briefly describe the VQGAN model with its codebook in this section, and more details can be found in~\cite{esser2021taming}.
VQGAN is comprised of an encoder $\encoder$, a decoder $\decoder$ and a codebook $\codebook=\{z_k\}^K_{k=1}$ with $K$ discrete codes.
For an input image $x\in \RR^{H\times W\times 3}$, the encoder $E$ maps the image $x$ to its spatial latent representation $\hat{z}=\encoder(x) \in \RR^{h\times w \times n_z}$, where $n_z$ is the dimension of latent vectors. 
The vector-quantized representation $\quantizedcode$ is then obtained by applying element-wise quantization $\quantize(\cdot)$ of each spatial code  $\hat{z}_{ij} \in \RR^{n_z}$ onto its closest codebook entry $z_k$:
\begin{equation}
  \quantizedcode = \quantize(\hat{z}) \coloneqq
  \left(\arg\min_{z_k \in \codebook} \Vert \hat{z}_{ij} - z_k \Vert\right)
  \in \RR^{h\times w \times n_z}.
  \label{eq:vq}
\end{equation}
The decoder $\decoder$ maps the quantized representation $\quantizedcode$ back to the image space, and the overall reconstruction $\hat{x} \approx x$ can be formulated as:
\begin{equation}
  \hat{x} = \decoder(\quantizedcode ) = \decoder\left(
    \quantize(\encoder(x)) \right).
\label{eq:vqrec}
\end{equation}

The encoder $E$ maps images of size $H\times W$ into discrete codes of size $H/f\times W/f$, where $f$ denotes the downsampling factor. 
It can be regarded as compressing each $f\times f$ patch in the image $x$ into one code.
In other words, for each code in $\quantizedcode$, $f$ also denotes the corresponding spatial size in the original image $x$.
We name this downsampling factor $f$ as the \textit{compression patch size} in our paper. 

Since the quantization operation in \eqref{eq:vqrec} is discrete and non-differentiable, VQGAN adopts straight-through gradient estimator~\cite{bengio2013estimating}, which simply copies the gradients from the decoder to the encoder.
Thus, the model and codebook can be trained end-to-end via the loss function $\mathcal{L}_{vq}$.
VQGAN also employs perceptual loss and adversarial loss to encourage reconstructions with better perceptual quality.
The full training objective of VQGAN and its codebook is:
\begin{equation}
  \mathcal{L}(\encoder, \decoder, \codebook) = \underbrace{\Vert x - \hat{x} \Vert_1 
  + \Vert \text{sg}[\encoder(x)] - \quantizedcode \Vert_2^2 + \beta\Vert \text{sg}[\quantizedcode] - \encoder(x) \Vert_2^2}_{\mathcal{L}_{vq}} + \mathcal{L}_{per} + \mathcal{L}_{adv},
  \label{eq:vqobjective}
\end{equation}
where $\Vert x - \hat{x}\Vert_1$ is the reconstruction loss and sg$[\cdot]$ denotes stop gradient operation.
The codebook is updated by $\Vert \text{sg}[\encoder(x)] - \quantizedcode \Vert_2^2$. $\beta\Vert \text{sg}[\quantizedcode] - \encoder(x) \Vert_2^2$ is the commitment loss~\cite{van2017neural} to reduce the discrepancy between the encoded latent vectors and codes, where $\beta$ is the commitment weight and set to 0.25 in all experiments.

\subsubsection{Analysis}
\label{vq_analysis}
In order to better understand the potential and limitations of VQ codebooks for face restoration, we conduct several preliminary experiments and draw the following observations.

\textbf{Observation 1}: \textit{Degradations in LQ faces can be removed by VQ codebooks trained only with HQ faces, 
when we adopt a proper compression patch size $f$.}

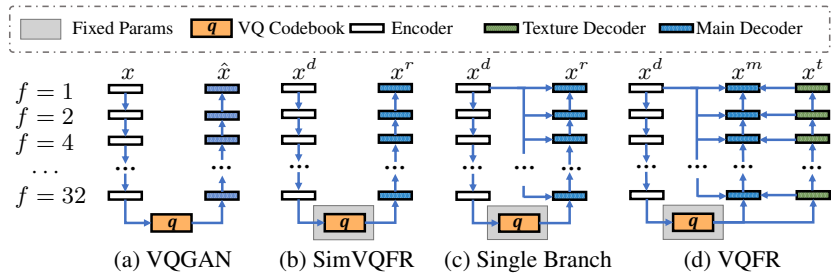
\begin{figure*}[!tb]
	\centering
	\begin{overpic}[width=0.9\textwidth]{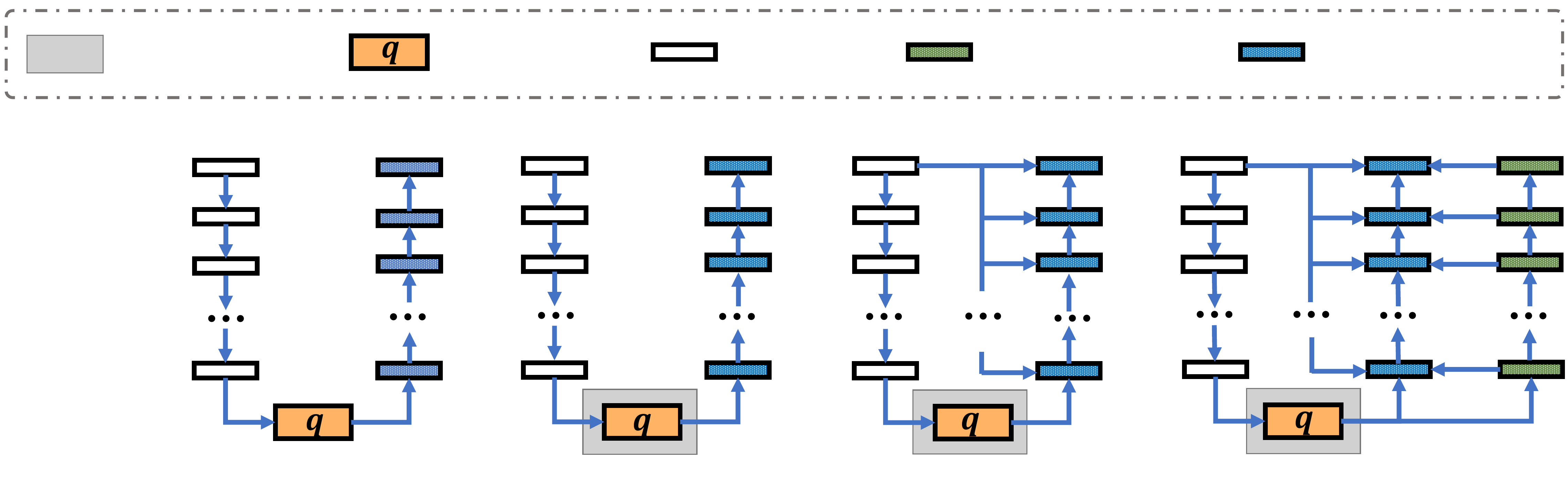}
        \put(8,27.3){\scriptsize Fixed Params}
        \put(28,27.3){\scriptsize VQ Codebook}
        \put(46.6,27.3){\scriptsize Encoder}
        \put(62.5,27.3){\scriptsize Texture Decoder}
        \put(83.5,27.3){\scriptsize Main Decoder}
        \put(13,-1){\footnotesize (a) VQGAN}
        \put(33,-1){\footnotesize (b) SimVQFR}
        \put(53,-1){\footnotesize (c) Single Branch}
        \put(82,-1){\footnotesize (d) VQFR}
        \put(13.9,21.8){\footnotesize $x$}
        \put(25.5,21.8){\footnotesize $\hat{x}$}
        \put(34.6,21.8){\footnotesize $x^d$}
        \put(46.5,21.8){\footnotesize $x^r$}
        \put(55.7,21.8){\footnotesize $x^d$}
        \put(67.6,21.8){\footnotesize $x^r$}
        \put(76.5,21.8){\footnotesize $x^d$}
        \put(87.8,21.8){\footnotesize $x^m$}
        \put(96.3,21.8){\footnotesize $x^t$}
        \put(1,19.4){\footnotesize $f=1$}
        \put(1,16.3){\footnotesize $f=2$}
        \put(1,13.2){\footnotesize $f=4$}
        \put(3,9.8){$\mathbf{\cdots}$}
        \put(1,6.8){\footnotesize $f=32$}
	\end{overpic}
	\caption{Illustration of the architecture variants. (a) The VQGAN structure is used in codebook training. (b) The SimVQFR structure. (c) Single branch decoder. (d) The proposed parallel decoder.}
	\label{fig:ParallelDec}
\end{figure*}

\noindent Following~\cite{esser2021taming} and \eqref{eq:vqobjective}, we first train VQGAN with perceptual codebooks on HQ faces using different compression patch sizes $f=\{8,16,32,64\}$. All our experiments are conducted on $512\times 512$ input faces. The illustration of training architectures is shown in \figref{fig:ParallelDec}.
We then examine the reconstruction quality for different compression patch sizes $f$.
Note that the reconstruction output is expected to be the same as the input.
As shown in row 1 of \figref{fig:motivation}, the reconstruction quality of HQ faces is as expected.
The reconstruction quality of HQ faces exhibits a reasonable trend: a smaller compression patch size $f$ will lead to more faithful outputs.
After that, we are curious whether the codebook trained only on HQ faces can also reconstruct LQ faces. We use LQ faces as inputs and examine the outputs. 
Interestingly, we can observe that under the compression patch sizes $f=32$, the degradation in LQ faces can be removed (see row 2 in \figref{fig:motivation}). This is because the codebook trained only on HQ faces has almost no degradation-related codes. During reconstruction, the vector quantization operation can replace the ``degraded" vectors of inputs with the ``clean" codes in codebooks.

Such a phenomenon shows the potential of the VQ codebook. However, it only happens with a large compression patch size $f$, as the codebook with a small compression patch size cannot well distinguish the degradations and detailed textures. 
In the extreme, for $f=1$ with each pixel having a quantized code, both the degradation and detailed textures can be well recovered by a combination of codebook entries.
On the other hand, a too large compression patch size (\textit{e.g.}, $f=64$) will result in a significant change in identity, even it could also reconstruct ``clean" face images.

\textbf{Observation 2}: \textit{When training for the restoration task, there is also a trade-off between improved detailed textures and fidelity changes.}

\noindent We then investigate the VQ codebooks in face restoration, \textit{i.e.}, training with LQ-HQ pairs as most face restoration works do~\cite{wang2021towards}. 
Based on the trained model for reconstruction, we then fix the VQ codebooks, and finetune the encoder and decoder with LQ-HQ pairs (the training details are the same as that in \secref{sec:implementation}). We denote this simple VQ model for face restoration as SimVQFR.
We can observe from \figref{fig:motivation} that there is still a trade-off between improved detailed textures and fidelity changes. The key influential factor is the compression patch size $f$.
With a small $f$ (\textit{i.e.}, $\{8,16\}$), the SimVQFR model fails to remove degradations and cannot recover sufficient detailed textures. While with a large $f$ (\textit{i.e.}, $\{32,64\}$), the textures are largely improved but the fidelity (\textit{i.e.}, expression and identity) also changes a lot by the codebook. 

\noindent\textbf{Our choice.}
Based on the above analysis, we can conclude that the VQ codebook, as a texture dictionary, has its value in generating high-quality facial textures and removing input degradations. 
However, there is a trade-off between the improved detailed textures and fidelity changes. The key influential factor is the compression patch size $f$.
In order to better leverage the strength of generating high-quality facial textures, we choose the compression patch size $f=32$ for $512\times 512$ input faces.
The left problem is how to preserve the fidelity with the VQ codebook of $f=32$. We will present our parallel decoder solution to address this problem in \secref{sec:parallel}.

\subsection{Parallel Decoder}
\label{sec:parallel}
The VQ codebook with the compression patch size of $f=32$ can be used as a texture bank to provide realistic facial details, but it also brings the problem of fidelity changes.
From the results in \figref{fig:motivation}, we can find that the position of facial components and the facial lines are changed, making the expression and identity largely deviate from the inputs.
A straightforward solution is to integrate the feature information from degraded inputs to help improve the fidelity.
However, simply fusing input features into the decoder with a single branch (as shown in \figref{fig:ParallelDec}) will lead to inferior details (see \figref{fig:visablation}).
In other words, such a single branch fusion strategy tends to corrupt the generated high-quality details.
Though input features can bring more fidelity information, these features also contain input degradations.
During feature propagation, \textit{i.e.}, the upsampling process from small spatial size to large spatial size, the intermediate features will largely be influenced by input features, and gradually contain fewer high-quality facial details from the VQ codebook.

To overcome the dilemma of keeping the realistic facial details and promoting fidelity, we propose a parallel decoder structure with a texture warping module.
The core idea of the parallel decoder (\figref{fig:ParallelDec}) is to decouple the two goals of face restoration, \textit{i.e.}, generating high-quality facial details and keeping the fidelity. 
As shown in \figref{fig:pipeline}, given a degraded face $x^{d}\in \RR^{H\times W\times 3}$, we first encode it to the latent vector $z^{d}$ by $z^{d}=\encoder(x^{d})$. The encoder consists of several residual blocks and downsampling operations. Then we replace the code $z^{d}$ with the HQ codebooks by \eqref{eq:vq} to get the quantized code $\quantizedcode^d$.
Since the quantized code $\quantizedcode^d$ is from the HQ codebook, it contains HQ facial details without degradations.
In order to keep its realistic textures, we use a texture decoder $\decoder_t$ to decode it back to image space $x^{t}=\decoder_t(\quantizedcode^d)$. We denote the multi-level feature of the texture branch as $F^t=\{F^t_{i}\}$. Since the texture branch only decodes from the HQ code, $F^t$ can keep realistic facial details. 

The main branch decoder $\decoder_m$ aims to generate faces $x^m$ with high fidelity while having the realistic facial details from the texture decoder $\decoder_t$.
As illustrated in \figref{fig:pipeline}, the main branch decoder $\decoder_m$ warps the texture feature $F^t$ based on the input feature extracted from degraded inputs at multiple spatial levels.
We use the input feature with the largest spatial resolution as it retains the richest fidelity information of degraded faces.
We then directly downsample the input feature to different resolution levels to get the multi-level features of degraded faces $F^d=\{F^d_{i}\}$.

For the $i$-th resolution level, we first warp $F^t_i$ with high-quality facial details towards $F^d_i$ by a texture warping module, which will be described later. 
After that, we fuse the warped feature $F^w_i$ and the upsampled feature from $F_{i-1}$, to get the $F_i$ feature in the main decoder.
The process can be formulated as:
\begin{equation}
    F^{w}_i=TWM(F^d_i,F^t_i), \qquad
    F_i=Conv(Concat(Upsample(F_{i-1}), F^w_i)),
\end{equation}
where $TWM$ is the texture warping module. 
Our parallel decoder shares the same spirits as reference-based restoration~\cite{li2018learning}. 
In our case, the features from the texture decoder serve as the reference features containing high-quality details.
Unlike reference-based restoration, our method does not require extra high-quality images with rich textures. Instead, the parallel decoder learns the main feature and ``reference" feature jointly.

\myPara{Texture Warping Module (TWM)}
The texture warping module aims to warp realistic facial details to match the fidelity of degraded inputs, especially the position of facial components and expressions. 
There are two inputs of TWM, one is the input feature $F^d$ and the other is the texture feature $F^t$.
As analyzed above, the texture features are decoded from the HQ codebook and have high-quality facial details. But their fidelity probably deviates from inputs.
Therefore, we adopt a deformable convolution~\cite{zhu2019deformable} to better warp the realistic facial details $F^t$ towards the input feature $F^d$.
Specifically, we first concatenate those two features to generate offsets. Then the offset is used in the deformable convolution to warp the texture features to match the fidelity of input, which can be formulated as:
\begin{equation}
    offset=Conv(Concat(F^d_i,F^t_i)),\qquad F^w_i=Dconv(F^t_i, offset),
\end{equation}
where $Dconv$ denotes the deformable convolution.
We also adopt a separable convolution with a large kernel size to model large position offsets between texture features and input features.

\subsection{Model Objective}
\label{sec:objective}
The training objective of VQFR consists of 1) pixel reconstruction loss that constraints the restored outputs close to the corresponding HQ faces; 2) code alignment loss that forces the codes of LQ inputs to match the codes of the corresponding HQ inputs; 3) perceptual loss to improve the perceptual quality in feature space; and 4) adversarial loss for restoring realistic textures. We denote the degraded face as $x^d$, decoder restored results as $x^r=\{x^t,x^m\}$ and the ground truth HQ image as $x^{h}$. The loss definitions are as follows.

\myPara{Pixel Reconstruction Loss.}
We use the widely-used L1 loss in the pixel space as the reconstruction loss, which is denoted as: $\mathcal{L}_{pix}=\Vert x^r - x^{h} \Vert_1$.
Empirically, we find that with the input feature of degraded faces, the pixel reconstruction loss has a negative impact on facial details. Therefore, we discard pixel reconstruction loss in the main decoder.

\myPara{Code Alignment Loss.}
The code alignment loss aims to improve the performance of matching the codes of LQ images with codes of HQ images.
We adopt the L2 loss to measure the distance, which can be formulated as:
$\mathcal{L}_{code}=\Vert z^{d} - \quantizedcode^{h} \Vert_2^2
$,
where $\quantizedcode^{h}$ is the ground truth code obtained from encoding HQ face to the quantized code by pretrained VQGAN.

\myPara{Perceptual Loss.}
We use the widely used perceptual loss~\cite{johnson2016perceptual,zhang2018perceptual} for both two decoder outputs: $  \mathcal{L}_{per}=\Vert \phi(x^r)-\phi(x^{h}) \Vert_1 + \lambda_{style}\Vert Gram(\phi(x^r))-Gram(\phi(x^{h})) \Vert_1$,
where $\phi$ is the pretrained VGG-16~\cite{simonyan2014very} network and $Gram$ means the Gram matrix~\cite{gatys2016image}. The former term measures the content distance and the latter term measures the style difference between the restoration results and the corresponding HQ images. 

\myPara{Adversarial Loss.}
We employ the global discriminator $D_g$ in SWAGAN~\cite{gal2021swagan} to encourage VQFR to favor the solutions in the natural image manifold and generate realistic textures.
In order to increase the local quality of facial details, we further adopt the local discriminator $D_l$ of PatchGAN~\cite{isola2017image} in our training. The objectives of the global and local discriminators are defined as 
\begin{equation}
\mathcal{L}^{global}_{adv} = -\mathbb{E}_{x^r} [\mathtt{softplus}(D_g(x^r))], 
\mathcal{L}^{local}_{adv} = \mathbb{E}_{x^r} [\log D_l(x^h) + \log (1 -D_l(x^r))].
\end{equation}

\myPara{Total objective.} The total training objective is the combination of above losses:
\begin{equation}
\mathcal{L}_{total} = \lambda_{pix}\mathcal{L}_{pix} + \lambda_{code}\mathcal{L}_{code} + \lambda_{per}\mathcal{L}_{per} +
\lambda_{global} \mathcal{L}^{global}_{adv} + \lambda_{local} \mathcal{L}^{local}_{adv},
\end{equation}
where the $\lambda_{pix}$, $\lambda_{code}$, $\lambda_{per}$, $\lambda_{global}$ and $\lambda_{local}$ are scale factors of corresponding loss.
\section{Experiments} \label{sec:experiments}

\begin{table}[!tb]
	\small
	\centering
	\caption{Quantitative comparison on the \textbf{CelebA-Test} dataset for blind face restoration. \redbf{Red} and \blueud{blue} indicates the best and second best performance.}
	\label{tab:celeba_blind}
	\tabcolsep=0.1cm
	\scalebox{0.95}{
		\begin{tabular}{c|ccc|cc|cc}
			\hline
			Methods       & LPIPS$\downarrow$  & FID$\downarrow$   &NIQE $\downarrow$    & Deg.$\downarrow$  & LMD.$\downarrow$   & PSNR$\uparrow$  & SSIM$\uparrow$  \\ \hline
			Input        & 0.4866 & 143.98  & 13.440 & 47.94 & 3.76 & 25.35  & 0.6848  \\
			Wan \etal.~\cite{wan2020bringing}  & 0.4826 & 67.58 & 5.356   & 43.00 & 2.92 & 24.71  & 0.6320  \\
			HiFaceGAN~\cite{yang2020hifacegan}      & 0.4770  & 66.09 & 4.916  & 42.18 & 3.16 & 24.92 & 0.6195  \\
			DFDNet~\cite{li2020blind}         & 0.4341 & 59.08 & 4.341   & 40.31 & 3.31 & 23.68  & 0.6622  \\
			PSFRGAN~\cite{chen2021progressive}       & 0.4240 & 47.59  & 5.123  & 39.69 & 3.41 & 24.71  & 0.6557  \\ \hline
			mGANprior~\cite{gu2020image}&   0.4584    &   82.27      &   6.422      &   55.45   & 411.72   &    24.30     &     0.6758   \\
			PULSE~\cite{menon2020pulse}&   0.4851     &   67.56      &   5.305      &   69.55   & 7.35   &    21.61    &     0.6200  \\
			GFP-GAN~\cite{wang2021towards} & \blueud{0.3646}  & \blueud{42.62}  & \blueud{4.077} & \redbf{34.60} &  \redbf{2.41} & 25.08   & 0.6777  \\ \hline
			\textbf{VQFR (ours)} & \redbf{0.3515}  & \redbf{41.28}  & \redbf{3.693}  & \blueud{35.75} & \blueud{2.43} & 24.14  & 0.6360 \\ \hline
	\end{tabular}}
\end{table}

\begin{figure*}[!tb]
	\centering
	\begin{overpic}[width=\textwidth]{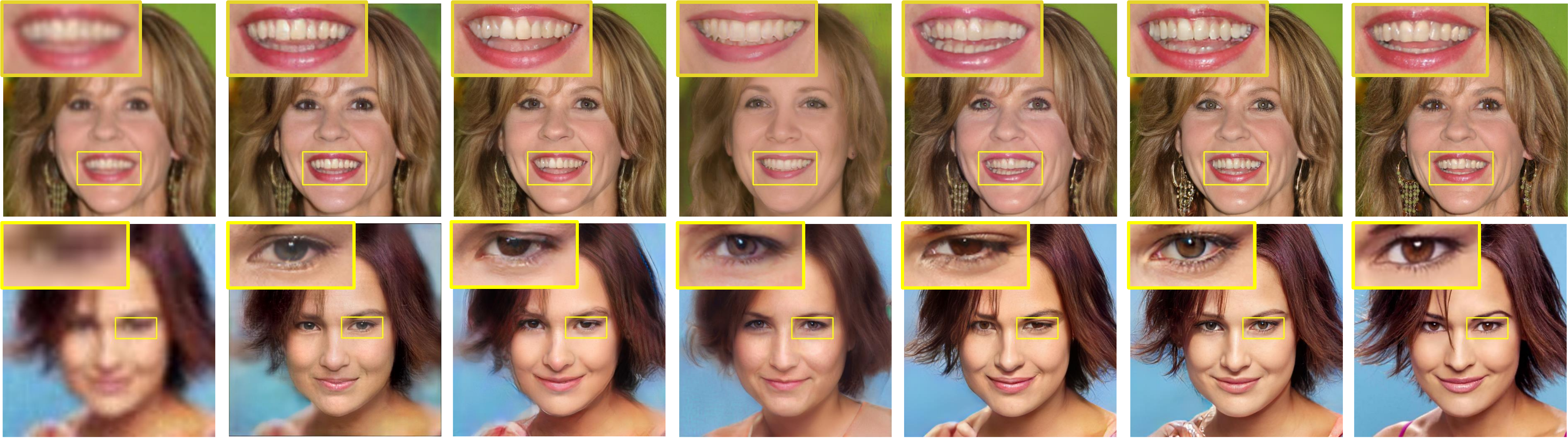}
		\put(5,-2){Input}
		\put(17,-2){DFDNet}
		\put(30,-2){PSFRGAN}
		\put(46,-2){PULSE}
		\put(59,-2){GFP-GAN}
		\put(75,-2){\textbf{VQFR}}
		\put(91.5,-2){GT}
	\end{overpic}
	\caption{Qualitative comparisons on the \textbf{CelebA-Test}. VQFR is able to restore high-quality facial details in various facial components, \textit{e.g.}, eyes and mouth.}
	\label{fig:syntheticcomp}
\end{figure*}
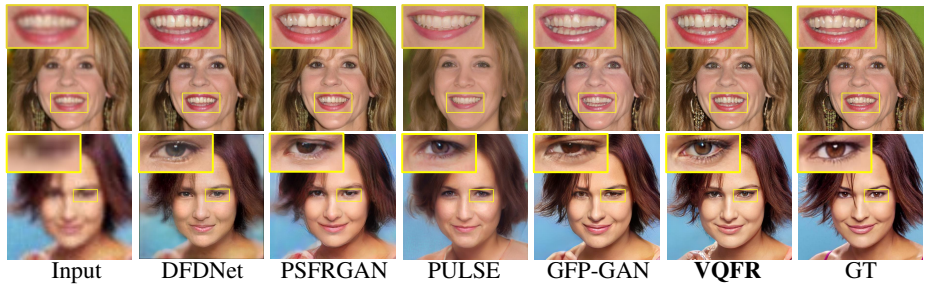

\subsection{Implementation and Evaluation Settings} 
\label{sec:implementation}

\myPara{Implementation.}
We implement the VQGAN and VQFR with six resolution levels, \textit{i.e.}, $\{1,2,4,8,16,32\}$. For the VQ codebook, we use 1024 codebook entries with 256 channels in all experiments. In order to increase the codebook usage, we follow~\cite{lancucki2020robust} and periodically re-initialize the codebook with k-means clustering.
More implementation details are provided \textit{in the \secref{sec:appendix}}.

In the first stage of codebook training, we adopt ADAM~\cite{kingma2014adam} optimizer with a learning rate of $1e-4$. The training iteration is set to 800K with a batch size of 16. For the second stage training for restoration, the loss weights are set to $\lambda_{pix}{=}\lambda_{code}{=}\lambda_{per}{=}1$, $ \lambda_{style}{=}2000$ and the adversarial loss weights are set to $\lambda_{local}{=}\lambda_{global}=0.5$. 
We train VQFR for 200K iterations by ADAM optimizer with the learning rate of $1e-5$. 

\myPara{Training Datasets.}
The VQFR is trained on the FFHQ dataset~\cite{karras2019style} including 70,000 high-quality faces. 
The images are resized to $512^2$ during training. 
Following the common practice in~\cite{li2020blind,li2018learning,wang2021towards}, we use the following degradation model to synthesize training pairs:
$x = [(y\circledast k_{\sigma})\downarrow_{r} + n_{\delta}]_{\mathtt{JPEG}_{q}}$,
where $\sigma$, $r$, $\delta$ and $q$ are randomly sampled from $\{0.2:10\}$, $\{1:8\}$, $\{0:15\}$ and $\{60:100\}$, respectively.

\begin{table}[!tb]
	\small
	\centering
	\caption{Quantitative comparison on the \textit{real-world} \textbf{LFW}, \textbf{CelebChild}, \textbf{WebPhoto}. \redbf{Red} and \blueud{blue} indicates the best and second best performance.}
	\label{tab:real_test}
	\tabcolsep=0.1cm
	\scalebox{0.95}{
		\begin{tabular}{c|cc|cc|cc}
			\hline
			Dataset       & \multicolumn{2}{c|}{\textbf{LFW-Test}} & \multicolumn{2}{c|}{\textbf{CelebChild}}   & \multicolumn{2}{c}{\textbf{WebPhoto}} \\ 
			Methods       & FID$\downarrow$  & NIQE $\downarrow$  & FID$\downarrow$  & NIQE $\downarrow$   &FID$\downarrow$  & NIQE $\downarrow$   \\ \hline
			Input         & 137.56 &11.214 & 144.42 & 9.170 & 170.11  & 12.755  \\
			Wan \etal.~\cite{wan2020bringing}  &   73.19    &   5.034     &   115.70     &   4.849     &    100.40     &     5.705   \\
			HiFaceGAN~\cite{yang2020hifacegan}     &   64.50    &   4.510     &   113.00     &   4.855     &    116.12    &     4.885  \\
			DFDNet~\cite{li2020blind}        &   62.57    &   4.026     &   111.55     &   4.414     &    100.68     &     5.293   \\
			PSFRGAN~\cite{chen2021progressive}      &   51.89    &   5.096     &   107.40     &   4.804     &    88.45     &     5.582  \\ \hline
			mGANprior~\cite{gu2020image}&   73.00    &   6.051     &   126.54     &   6.841      &    120.75     &     7.226   \\
			PULSE~\cite{menon2020pulse}&   64.86     &   5.097      &   \redbf{102.74}      &   5.225      &   86.45     &     5.146  \\
			GFP-GAN~\cite{wang2021towards} &   \redbf{49.96}     &   \blueud{3.882}    &   111.78     &     \blueud{4.349}   &   \blueud{87.35}     &     \blueud{4.144}    \\ \hline
			\textbf{VQFR (ours)} &  \blueud{50.64}    &   \redbf{3.589}   &   \blueud{105.18}     &  \redbf{3.936} &   \redbf{75.38}     &    \redbf{3.607}    \\ \hline
	\end{tabular}}
\end{table}

\myPara{Testing Datasets.}
Following the practice in GFP-GAN~\cite{wang2021towards}, we conduct experiments on the synthetic dataset~\textit{CelebA-Test} and three real-world datasets - \textit{LFW-Test}, \textit{CelebChild-Test} and \textit{WebPhoto-Test}. 
These datasets have diverse and complicated degradations.
All these datasets have no overlap with the training dataset.

\myPara{Evaluation Metrics.}\label{meric}
Our evaluation metrics contain two widely-used non-reference perceptual metrics: FID~\cite{heusel2017gans} and NIQE~\cite{mittal2012making}. 
We also measure the pixel-wise metrics (PSNR and SSIM) and perceptual metric (LPIPS~\cite{zhang2018unreasonable}) for benchmarking CelebA-Test with Ground-Truth (GT). 
We follow previous work~\cite{wang2021towards} to use the embedding angle of ArcFace~\cite{deng2019arcface} as the identity metric, which is denoted by `Deg.'. 
In order to better measure the fidelity with accurate facial positions and expressions, we further adopt landmark distance (LMD) as the fidelity metric. More details are provided \textit{in the \secref{sec:appendix}}.

\subsection{Comparisons with State-of-the-art Methods} 
We compare our VQFR with several state-of-the-art face restoration methods: Wan \etal.~\cite{wan2020bringing}, HiFaceGAN~\cite{yang2020hifacegan}, DFDNet~\cite{li2020blind}, PSFRGAN~\cite{chen2021progressive}, mGANprior~\cite{gu2020image}, PULSE~\cite{menon2020pulse} and GFP-GAN~\cite{wang2021towards}.

\myPara{Synthetic CelebA-Test.}
From the quantitative results in \tabref{tab:celeba_blind}, VQFR achieves the lowest LPIPS, implying that VQFR generates the restored faces with the closest perceptual quality to ground-truth.
VQFR also achieves the best FID and NIQE, with a large improvement over GFP-GAN, demonstrating that results of VQFR are closer to real faces and  have more realistic details.
For fidelity, VQFR can achieve comparable landmark distance and identity degree to GFP-GAN, showing that it can recover accurate facial expressions and detail positions. 

Qualitative results are presented in \figref{fig:syntheticcomp}. Thanks to the VQ codebook design, VQFR generates high-quality facial components like eyes and mouth as well as other facial regions.%
VQFR also maintains the fidelity with the help of the parallel decoder.

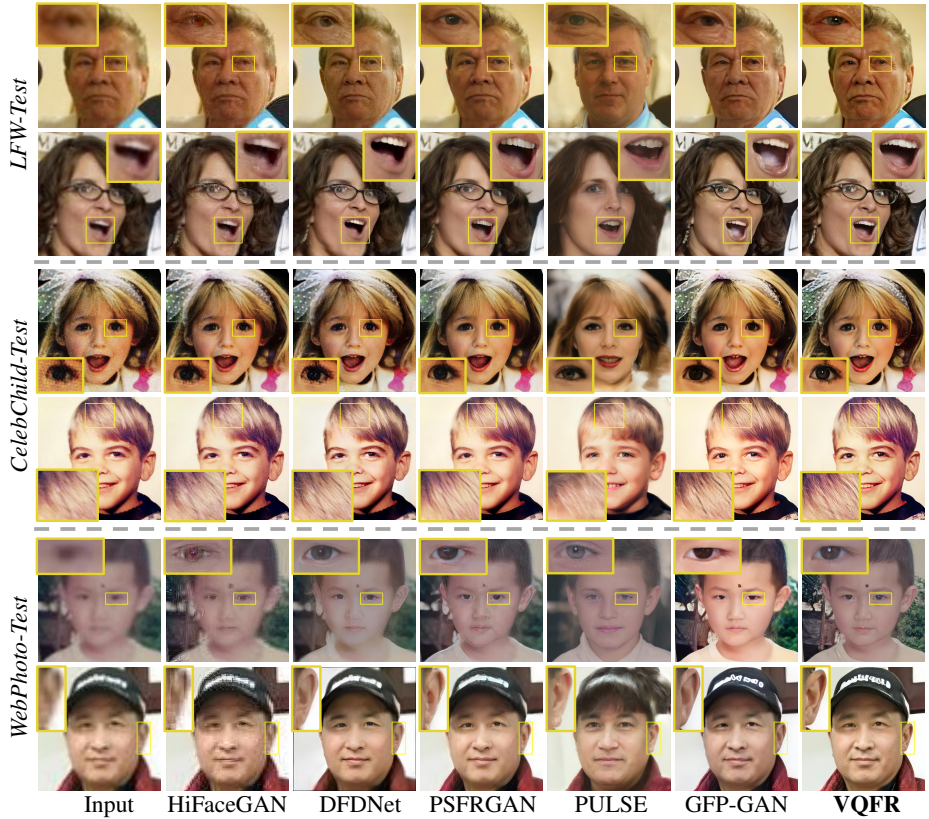
\begin{figure*}[!tb]
	\centering
	\begin{overpic}[width=\textwidth]{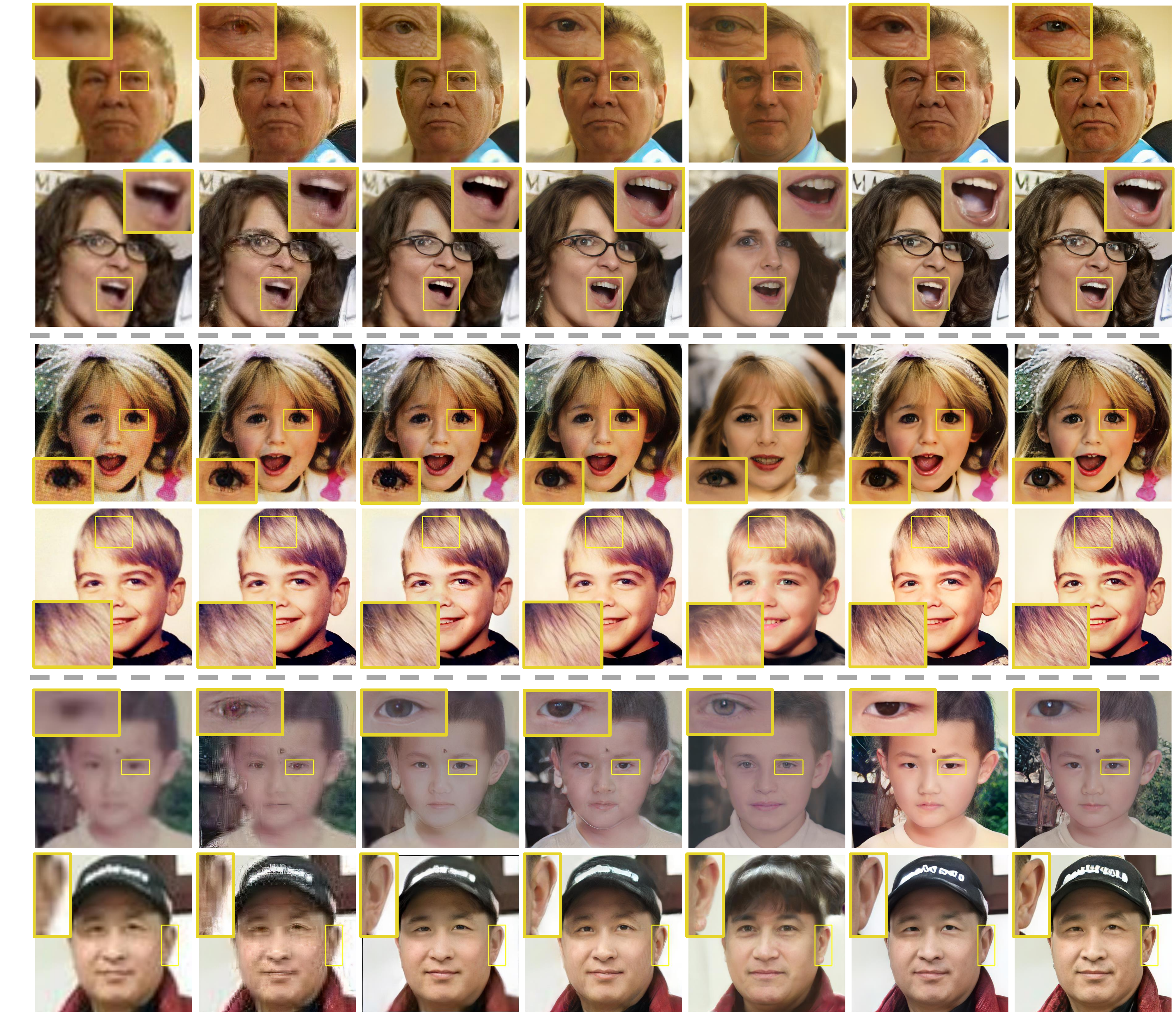}
		\put(0,68){\rotatebox{90}{\textit{LFW-Test}}}
		\put(0,35){\rotatebox{90}{\textit{CelebChild-Test}}}
		\put(0,6){\rotatebox{90}{\textit{WebPhoto-Test}}}

		\put(8,-2){Input}
		\put(17.1,-2){HiFaceGAN}
		\put(33,-2){ DFDNet}
		\put(45,-2){ PSFRGAN}
		\put(61.5,-2){PULSE}
		\put(73.5,-2){GFP-GAN}
		\put(89,-2){ \textbf{VQFR}}
	\end{overpic}
	\caption{Qualitative comparisons on three real-world datasets. \textbf{Zoom in for best view.}}
	\label{fig:realcomp}
\end{figure*}

\myPara{Real-World LFW, CelebChild, and WedPhoto-Test.}
We evaluate VQFR on three real-world test datasets to test the generalization ability. \tabref{tab:real_test} shows the quantitative results. 
It is observed that VQFR largely improves the realness and perceptual quality of all three real-world datasets. 
PULSE~\cite{menon2020pulse} achieves a higher perceptual quality on CelebChild, but its fidelity is severely affected. 
From the qualitative results in \figref{fig:realcomp}, the face restored by VQFR is of the most high quality on different facial regions.

\subsection{Ablation Study}
\begin{figure*}[!tb]
	\begin{minipage}[h]{0.5\linewidth}
	    \centering
		\subfloat[Ablation results on CelebA-Test. \textit{Inp Feat.}: input feature of degraded faces; \textit{Para Dec.}: parallel decoder; \textit{TWM}: texture warping module.]{
		\scalebox{0.75}{
		    \begin{tabular}{l|ccc|ccc}
        			\hline
	         \multirow{2}{*}{\textbf{Models}} & \multicolumn{3}{c|}{\textbf{Configurations}}  & \multicolumn{3}{c}{\textbf{CelebA-Test}} \\
        			 & Inp Feat.       & Para Dec.      & TWM       &  FID$\downarrow$  &  NIQE$\downarrow$  & LMD $\downarrow$   \\ \hline
        			SimVQFR &  &&&      40.51   &    3.844      &   2.96   \\\hline
        			Variant 1  & \checkmark & & & 39.64  &  4.019 &    2.36 \\
        			Variant 2  & \checkmark & \checkmark &&   42.61  &  3.714  &   2.48 \\\hline
        		    VQFR & \checkmark & \checkmark & \checkmark &    41.28  &   3.693     &    2.43    \\\hline
        	\end{tabular}}
		}\\
	 \subfloat[Visualization of the balance between realness of fidelity of different configurations.]{
		\includegraphics[width=0.98\textwidth]{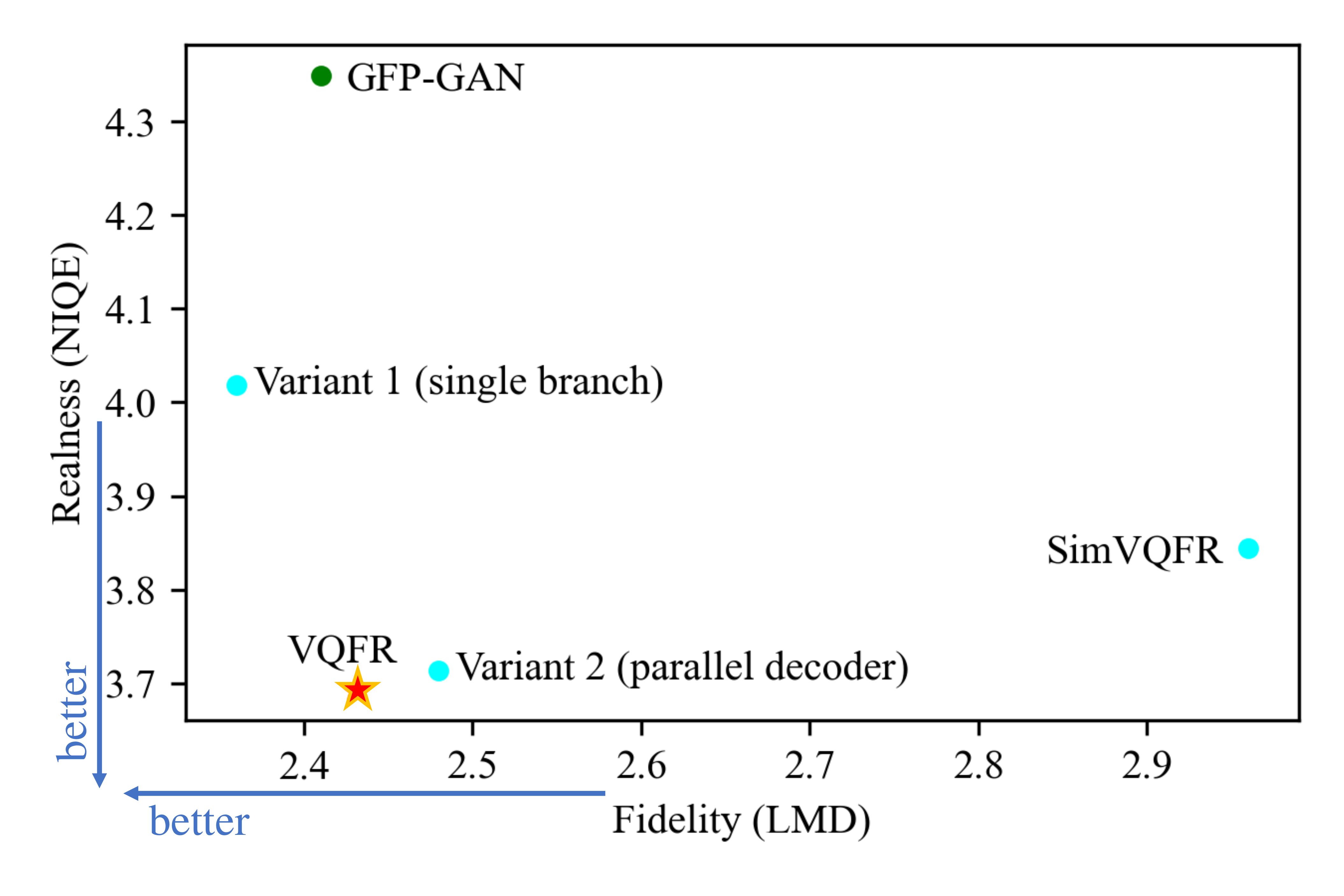}
		}
	\captionof{figure}{Quantitative ablation studies of key designs in VQFR.}
	\label{fig:ablation}
	\end{minipage}
	\hfill
	\begin{minipage}[h]{0.45\linewidth}
		\includegraphics[width=\textwidth]{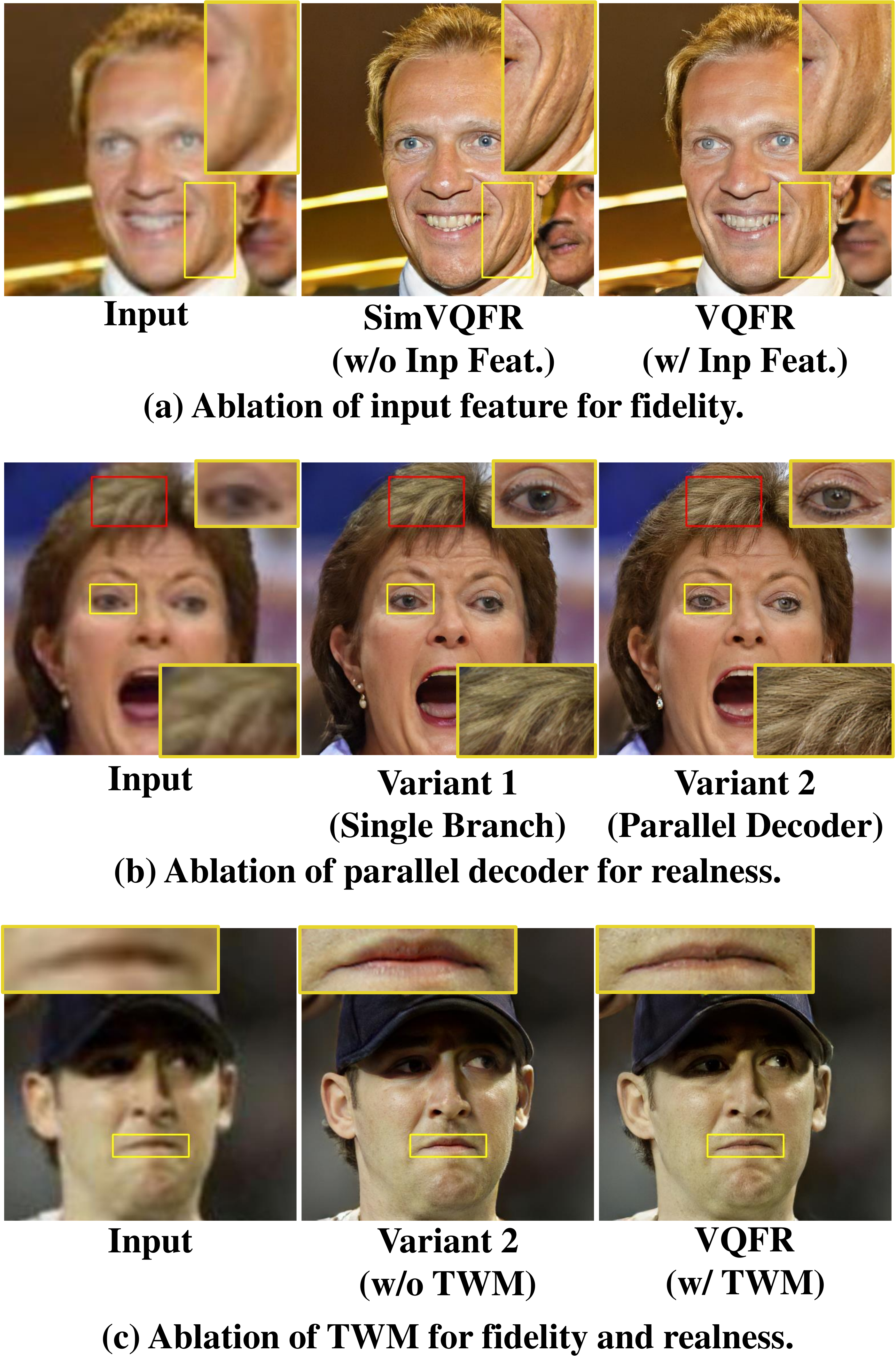}
		\captionof{figure}{Qualitative ablation studies of key designs on VQFR.}
		\label{fig:visablation}
	\end{minipage}
\end{figure*}

\myPara{Importance of input features from degraded faces.}
The SimVQFR model directly uses the VQ codebook without exploiting input features. 
As shown in Fig.~\ref{fig:ablation}(a), 
though it can achieve high perceptual quality (low FID and NIQE), its landmark distance is large, indicating the low fidelity. 
After incorporating the input features of degraded faces, the fidelity improves significantly. From the visualization of \figref{fig:visablation}(a), we can observe that the facial lines of SimVQFR are deviated from LQ faces, resulting in an expression change.
Such a phenomenon can also be observed in the examples in Fig.~\ref{fig:pipeline}, where the texture decoder output changes the woman's expression apparently. 

\myPara{Importance of parallel decoder.}
When comparing Variant 1 and Variant 2, the NIQE (which favors high-quality details) of Variants 1 is higher and clearly inferior to Variant 2 (Fig.~\ref{fig:ablation} (a)). 
From the visualization in \figref{fig:visablation}(b), we observe that, in  Variant 1 without the parallel decoder, the eyes and hair lose high-frequency details and are biased to degraded ones. While with the parallel decoder, the details are clearer and more realistic.

\myPara{Texture Warping Module.}
As shown in Fig.~\ref{fig:ablation} (a), Variant 2 does not have a TWM module and directly utilizes concatenation fusion of texture features and input features of degraded features. 
It lacks the ability to dynamically adjust the features with warping.
Therefore, Variant 2 without TWM keeps the high-quality textures but changes the fidelity, as it cannot well adjust the fine details and expression.
From the visualization of \figref{fig:visablation}(c), the TWM module helps accurately warp the high-quality facial details to better match the degraded input.

\section{Conclusion}
In this paper, we propose the vector-quantized (VQ) dictionary of high-quality facial details for face restoration. Our analysis of the VQ codebook shows the potential and limitations of the VQ codebook.
In order to keep the fidelity without the loss of high-quality facial details, 
a parallel decoder is further proposed to gradually fuse input features and texture features from VQ codebooks.
Equipped with the VQ codebook as a dictionary and the parallel decoder, our proposed vector-quantized face restoration (VQFR) can produce high-quality facial details while preserving fidelity.
Extensive experiments show that our methods surpass previous works on both synthetic and real-world datasets.
\section*{Acknowledgement}

This work is funded by the National Key Research and Development Program 
of China Grant No.2018AAA0100400 and NSFC (NO. 62176130). And this work is partially supported by National Natural Science Foundation of China (61906184, U1913210), and the Shanghai Committee of Science and Technology, China (Grant No. 21DZ1100100).

\bibliographystyle{splncs04}
\bibliography{VQFR}

\clearpage

\section{Appendix}
\label{sec:appendix}

In this section, we first present the \textbf{architecture details of VQFR} in \secref{sec:network}. Then we give more details about \textbf{evaluation metrics} in \secref{sec:metric}. The \textbf{limitations} of VQFR are discussed in \secref{sec:limitation}.
We then provide more \textbf{visual comparisons of ablation studies} to help better understand the VQFR designs in \secref{sec:abaltion}. More \textbf{visual comparisons with previous methods on the real-world datasets} are shown in \secref{sec:viscompare}. 

\subsection{Network Architectures}
\label{sec:network}
\myPara{VQFR:} The detailed architecture of VQFR is illustrated in \tabref{tab:arch_detail}. There are six resolution levels, \textit{i.e.}, $f=\{1,2,4,8,16,32\}$, and the quantization operation is conducted on the feature level of $f32$. Each level of the encoder contains two residual blocks, and each level of the texture branch in the decoder contains three residual blocks. Each level of the main branch in the decoder contains one texture warping module and one residual block. We use a bilinear upsample/downsample followed by a $1{\times}1$ convolution to change the resolutions. VQFR has 76.3M params (1.07 TFlops) and takes 0.36s to process a $512^2$ image on Nvidia A100.

\begin{table}[h]
    \small
    \centering
    \caption{The detailed architecture of VQFR. The residual block consists of $3{\times}3$ Conv-GN~\cite{wu2018group}-Swish~\cite{ramachandran2017searching}-$3{\times}3$ Conv-GN-Swish. g: groups in GroupNorm (GN); c: channels; dg: deformable groups in deformable convolution~\cite{zhu2019deformable}; f: compression patch size.}
    \label{tab:arch_detail}
    \resizebox{.98\textwidth}{!}{
    \begin{tabular}{c | c | c | c}
    \toprule
        Input size & Encoder & Texture branch & Main branch \\
    \midrule
        \multirow{3}*{$f1:$ 512$\times$512} & \makecell[c]{$\left\{
                        \begin{array}{c}
                        \text{Residual block, 128-c, 32-g}
                        \end{array}\right\} \times 2$} & \multirow{3}*{\makecell[c]{$\left\{
                        \begin{array}{c}
                        \text{Residual block, 128-c, 32-g} \\
                        \end{array}\right\} \times 3$}} & \multirow{3}*{\makecell[c]{$
			            \begin{array}{c}
			           	\text{TWM, 128-c, 4-dg}\\
			            \left\{\text{Residual block, 128-c, 32-g}\right\} \times 1
			        	\end{array}$}}\\ \cline{2-2}     
     ~ & \makecell[c]{$
                        \begin{array}{c}
                        \text{Bilinear downsampling 2$\times$} \\
                        \text{Conv $1\times 1$, 128-c}
                        \end{array}$} & ~ & ~\\
    \midrule
        \multirow{3}*{$f2:$ 256$\times$256} & \makecell[c]{$\left\{
        	\begin{array}{c}
        		\text{Residual block, 128-c, 32-g}
        	\end{array}\right\} \times 2$} & \makecell[c]{$\left\{
		        \begin{array}{c}
		        \text{Residual block, 128-c, 32-g} \\
		    \end{array}\right\} \times 3$} & \multirow{3}*{\makecell[c]{$
        		\begin{array}{c}
        	       \text{TWM, 128-c, 4-dg}\\
        			\left\{\text{Residual block, 128-c, 32-g}\right\} \times 1
        		\end{array}$}}\\ \cline{2-2} \cline{3-3}  
        	   
		        ~ & \makecell[c]{$
		        	\begin{array}{c}
		        		\text{Bilinear downsampling 2$\times$} \\
		        		\text{Conv $1\times 1$, 128-c  $\rightarrow$ 256-c}
		        	\end{array}$} & \makecell[c]{$
		    \begin{array}{c}
		        \text{Bilinear upsampling 2$\times$} \\
		        \text{Conv $1\times 1$, 128-c}
		    \end{array}$} & ~\\   
	\midrule
		\multirow{3}*{$f4:$ 128$\times$128} & \makecell[c]{$\left\{
			\begin{array}{c}
				\text{Residual block, 256-c, 32-g}
			\end{array}\right\} \times 2$} & \makecell[c]{$\left\{
			\begin{array}{c}
				\text{Residual block, 256-c, 32-g} \\
			\end{array}\right\} \times 3$} & \multirow{3}*{\makecell[c]{$
				\begin{array}{c}
					\text{TWM, 256-c, 4-dg}\\
					\left\{\text{Residual block, 256-c, 32-g}\right\} \times 1
				\end{array}$}}\\ \cline{2-2} \cline{3-3}
			    
		~ & \makecell[c]{$
			\begin{array}{c}
				\text{Bilinear downsampling 2$\times$} \\
				\text{Conv $1\times 1$, 256-c}
			\end{array}$} & \makecell[c]{$
			\begin{array}{c}
				\text{Bilinear upsampling 2$\times$} \\
				\text{Conv $1\times 1$, 256-c $\rightarrow$ 128-c}
			\end{array}$} & ~\\     
	\midrule
	\multirow{3}*{$f8:$ 64$\times$64} & \makecell[c]{$\left\{
		\begin{array}{c}
			\text{Residual block, 256-c, 32-g}
		\end{array}\right\} \times 2$} & \makecell[c]{$\left\{
		\begin{array}{c}
			\text{Residual block, 256-c, 32-g} \\
		\end{array}\right\} \times 3$} & \multirow{3}*{\makecell[c]{$
			\begin{array}{c}
				\text{TWM, 256-c, 4-dg}\\
				\left\{\text{Residual block, 256-c, 32-g}\right\} \times 1
			\end{array}$}}\\ \cline{2-2} \cline{3-3}
	
	~ & \makecell[c]{$
		\begin{array}{c}
			\text{Bilinear downsampling 2$\times$} \\
			\text{Conv $1\times 1$, 256-c}
		\end{array}$} & \makecell[c]{$
		\begin{array}{c}
			\text{Bilinear upsampling 2$\times$} \\
			\text{Conv $1\times 1$, 256-c}
		\end{array}$} & ~\\
	\midrule
	\multirow{3}*{$f16:$ 32$\times$32} & \makecell[c]{$\left\{
		\begin{array}{c}
			\text{Residual block, 256-c, 32-g}
		\end{array}\right\} \times 2$} & \makecell[c]{$\left\{
		\begin{array}{c}
			\text{Residual block, 256-c, 32-g} \\
		\end{array}\right\} \times 3$} & \multirow{3}*{\makecell[c]{$
			\begin{array}{c}
				\text{TWM, 256-c, 4-dg}\\
				\left\{\text{Residual block, 256-c, 32-g}\right\} \times 1
			\end{array}$}}\\ \cline{2-2} \cline{3-3}
	
	~ & \makecell[c]{$
		\begin{array}{c}
			\text{Bilinear downsampling 2$\times$} \\
			\text{Conv $1\times 1$, 256-c $\rightarrow$ 512-c}
		\end{array}$} & \makecell[c]{$
		\begin{array}{c}
			\text{Bilinear upsampling 2$\times$} \\
			\text{Conv $1\times 1$, 256-c}
		\end{array}$} & ~\\
	\midrule
	\multirow{3}*{$f32:$ 16$\times$16} & \multirow{3}*{\makecell[c]{$\left\{
		\begin{array}{c}
			\text{Residual block, 512-c, 32-g}
		\end{array}\right\} \times 2$}} & \makecell[c]{$\left\{
		\begin{array}{c}
			\text{Residual block, 512-c, 32-g} \\
		\end{array}\right\} \times 3$} & \multirow{3}*{\makecell[c]{$
			\begin{array}{c}				
			    \text{TWM, 512-c, 4-dg}\\
				\left\{\text{Residual block, 512-c, 32-g}\right\} \times 1
			\end{array}$}}\\\cline{3-3}
	
	~ & ~ & \makecell[c]{$
		\begin{array}{c}
			\text{Bilinear upsampling 2$\times$} \\
			\text{Conv $1\times 1$, 512-c $\rightarrow$ 256-c}
		\end{array}$} & ~\\   
    \bottomrule
    \end{tabular}}
\end{table}

\begin{table}[h]
    \small
    \centering
    \caption{The detailed architecture of the texture warping module (TWM). OConv: convolution for generating offsets; DConv: deformable convolution; c: channels; dg: deformable groups.}
    \label{tab:twm_detail}
    \resizebox{.98\textwidth}{!}{
    \begin{tabular}{c | c | c | c}
    \toprule
        Input size & $f1:$ 512$\times$512 & $f2:$ 256$\times$256 & $f4:$ 128$\times$128 \\
    \midrule
        \multirow{2}*{TWM} & \makecell[c]{OConv:$\left\{
                        \begin{array}{c}
                        \text{Conv 1$\times$1, (128+32)-c $\rightarrow$ 128-c}\\
                        \text{Depthwise Conv 7$\times$7, 128-c}\\
                        \text{Conv 1$\times$1, 128-c}
                    \end{array}\right\}$} & \makecell[c]{OConv:$\left\{
                        \begin{array}{c}
                        \text{Conv 1$\times$1, (128+32)-c $\rightarrow$ 128-c}\\
                        \text{Depthwise Conv 7$\times$7, 128-c}\\
                        \text{Conv 1$\times$1, 256-c}
                        \end{array}\right\}$} & \makecell[c]{OConv:$\left\{
                        \begin{array}{c}
                        \text{Conv 1$\times$1, (256+32)-c $\rightarrow$ 256-c}\\
                        \text{Depthwise Conv 7$\times$7, 256-c}\\
                        \text{Conv 1$\times$1, 256-c}
                        \end{array}\right\}$} \\ \cline{2-4}     
     ~ & \makecell[c]{DConv:$\left\{
                        \begin{array}{c}
                        \text{Deformable Conv 3$\times$3, 128-c, 4-dg}
                        \end{array}\right\}$} & \makecell[c]{DConv:$\left\{
                        \begin{array}{c}
                        \text{Deformable Conv 3$\times$3, 128-c, 4-dg}
                        \end{array}\right\}$} & \makecell[c]{DConv:$\left\{
                        \begin{array}{c}
                        \text{Deformable Conv 3$\times$3, 256-c, 4-dg}
                        \end{array}\right\}$}\\
    \bottomrule
    \end{tabular}}
    \resizebox{.98\textwidth}{!}{
    \begin{tabular}{c | c | c | c}
    \toprule
        Input size & $f8:$ 64$\times$64 & $f16:$ 32$\times$32 & $f32:$ 16$\times$16 \\
    \midrule
        \multirow{2}*{TWM} & \makecell[c]{OConv:$\left\{
                        \begin{array}{c}
                        \text{Conv 1$\times$1, (256+32)-c $\rightarrow$ 256-c}\\
                        \text{Depthwise Conv 7$\times$7, 256-c}\\
                        \text{Conv 1$\times$1, 256-c}
                    \end{array}\right\}$} & \makecell[c]{OConv:$\left\{
                        \begin{array}{c}
                        \text{Conv 1$\times$1, (256+32)-c $\rightarrow$ 256-c}\\
                        \text{Depthwise Conv 7$\times$7, 256-c}\\
                        \text{Conv 1$\times$1, 256-c}
                        \end{array}\right\}$} & \makecell[c]{OConv:$\left\{
                        \begin{array}{c}
                        \text{Conv 1$\times$1, (512+32)-c $\rightarrow$ 512-c}\\
                        \text{Depthwise Conv 7$\times$7, 512-c}\\
                        \text{Conv 1$\times$1, 512-c}
                        \end{array}\right\}$} \\ \cline{2-4}     
     ~ & \makecell[c]{DConv:$\left\{
                        \begin{array}{c}
                        \text{Deformable Conv 3$\times$3, 256-c, 4-dg}
                        \end{array}\right\}$} & \makecell[c]{DConv:$\left\{
                        \begin{array}{c}
                        \text{Deformable Conv 3$\times$3, 256-c, 4-dg}
                        \end{array}\right\}$} & \makecell[c]{DConv:$\left\{
                        \begin{array}{c}
                        \text{Deformable Conv 3$\times$3, 512-c, 4-dg}
                        \end{array}\right\}$}\\
    \bottomrule
    \end{tabular}}
\end{table}

\myPara{Texture Warping Module (TWM):}
We use a $3\times3$ convolution with 32 output channels to extract input information of degraded faces. Then we resize the feature to match all resolution levels ($f={1,2,4,8,16,32}$). The detailed architecture of TWM is shown in \tabref{tab:twm_detail}. For each resolution level, the offset convolution is used to generate offsets from the concatenation of the texture feature and the input features of degraded faces. Then, the offsets and the texture features are fed into the deformable convolution, outputting the warped feature.

\subsection{Evaluation Metrics}
\label{sec:metric}
Our evaluation metrics contain two widely-used non-reference perceptual metrics: FID~\cite{heusel2017gans} and NIQE~\cite{mittal2012making}. 
We also measure the pixel-wise metrics (PSNR and SSIM) and perceptual metric (LPIPS~\cite{zhang2018unreasonable}) for benchmarking CelebA-Test with Ground-Truth (GT). 
However, as pointed out in \cite{blau2018perception}, the distortion measure (\textit{e.g.}, PSNR, SSIM) and perceptual quality are at odds with each other.
Similar to GFP-GAN~\cite{wang2021towards}, we pursue perceptual quality in VQFR and provide PSNR/SSIM for reference only.
In the Table. 1 of the main manuscript, the best PSNR and SSIM are achieved by degraded inputs, as all other methods are optimized for the perceptual quality instead of the distortion measures.

For fidelity measurement, we follow previous work~\cite{wang2021towards} to use the embedding angle of ArcFace~\cite{deng2019arcface} as the identity metric, which is denoted by `Deg.'.
However, this Deg. metric actually cannot well reflect the fidelity due to the following reasons. 1) The ArcFace model downsamples the face images into $128{\times}128$ during inference, which loses the spatial dimension. Thus, it cannot evaluate detailed  facial positions. 2) The ArcFace is designed for the recognition task and is trained with the invariance to expressions. 
While the expression is important for measuring fidelity in face restoration.
In order to better measure the fidelity with accurate detailed facial positions and expressions, we further adopt landmark distance (LMD) as the fidelity metric. Specially, we use AWing~\cite{wang2019adaptive} to obtain 98 landmarks for both the restored face and the ground-truth face. Then we calculate the L2 distance for each landmark and average the distance as the final score of the LMD metric.

\begin{figure*}[!tb]
	\centering
	\includegraphics[width=\linewidth]{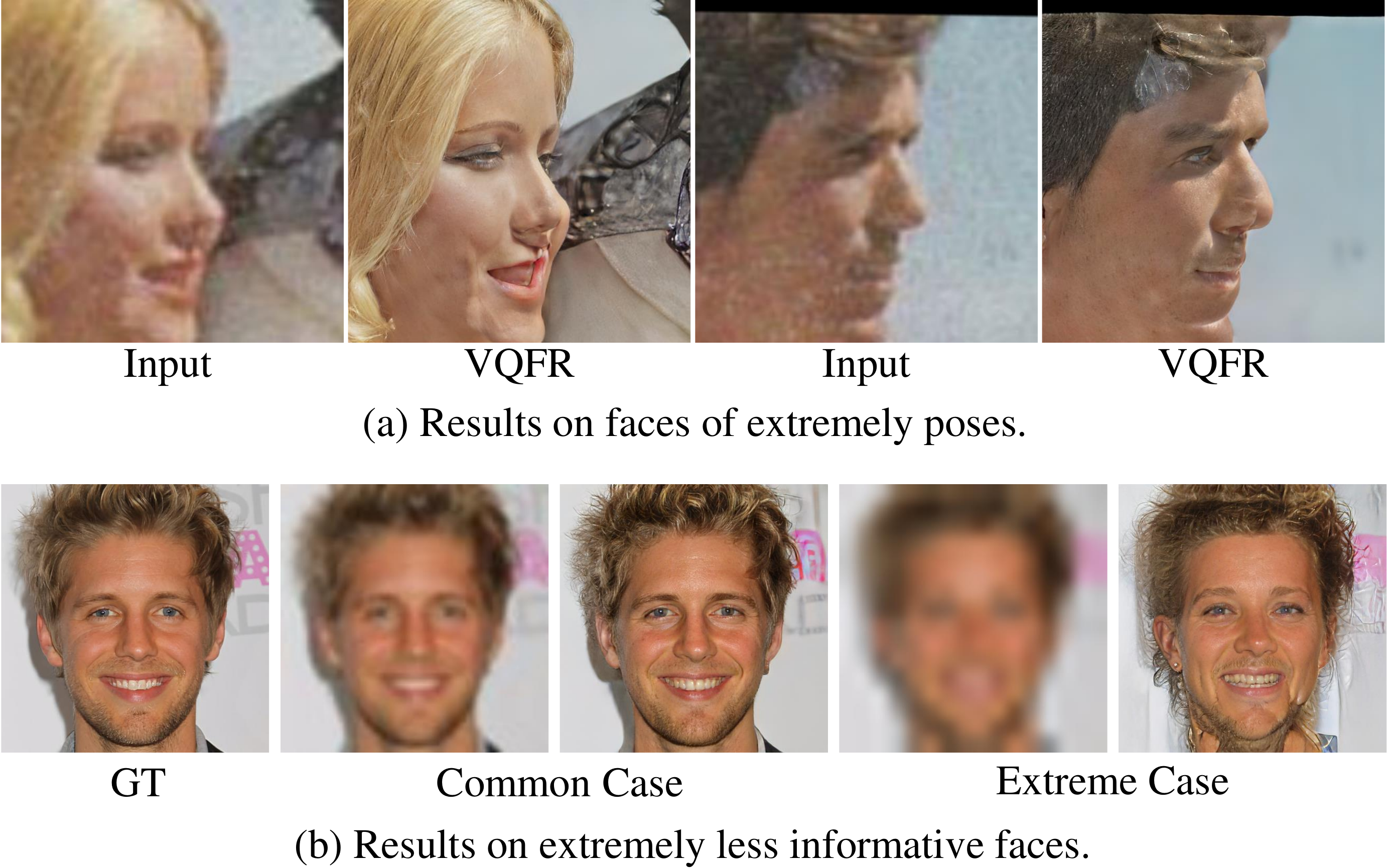}
	\caption{Limitations of VQFR.}
	\label{fig:limitation}
\end{figure*}

\subsection{Limitation}
\label{sec:limitation}
The limitations of VQFR are two-folds. 1) As shown in \figref{fig:limitation}(a), faces in extreme poses
lead to poor restoration results, since the codebook is built from
the training dataset, in which most samples are frontal faces.
One potential solution is to increase the dataset diversity and codebook size, which will help
build a more comprehensive dictionary. 2) As shown in \figref{fig:limitation}(b),
the restoration from extremely less informative faces is far from
satisfactory, since the VQFR does not build upon a generative
model. Moreover, VQ may further lead to divergent codebook
quantization due to the less informative inputs.
One promising direction to improve is to equip VQFR with generation ability. For example, when the input faces contain extreme low-information and thus the code mapping is ambiguous, the auto-regressive~\cite{esser2021taming} or bi-directional~\cite{devlin2018bert} transformer can help model the code selection.

\subsection{More Visualizations of Ablation Study}
\label{sec:abaltion}

\begin{figure*}[!tb]
	\centering
	\begin{overpic}[width=\linewidth]{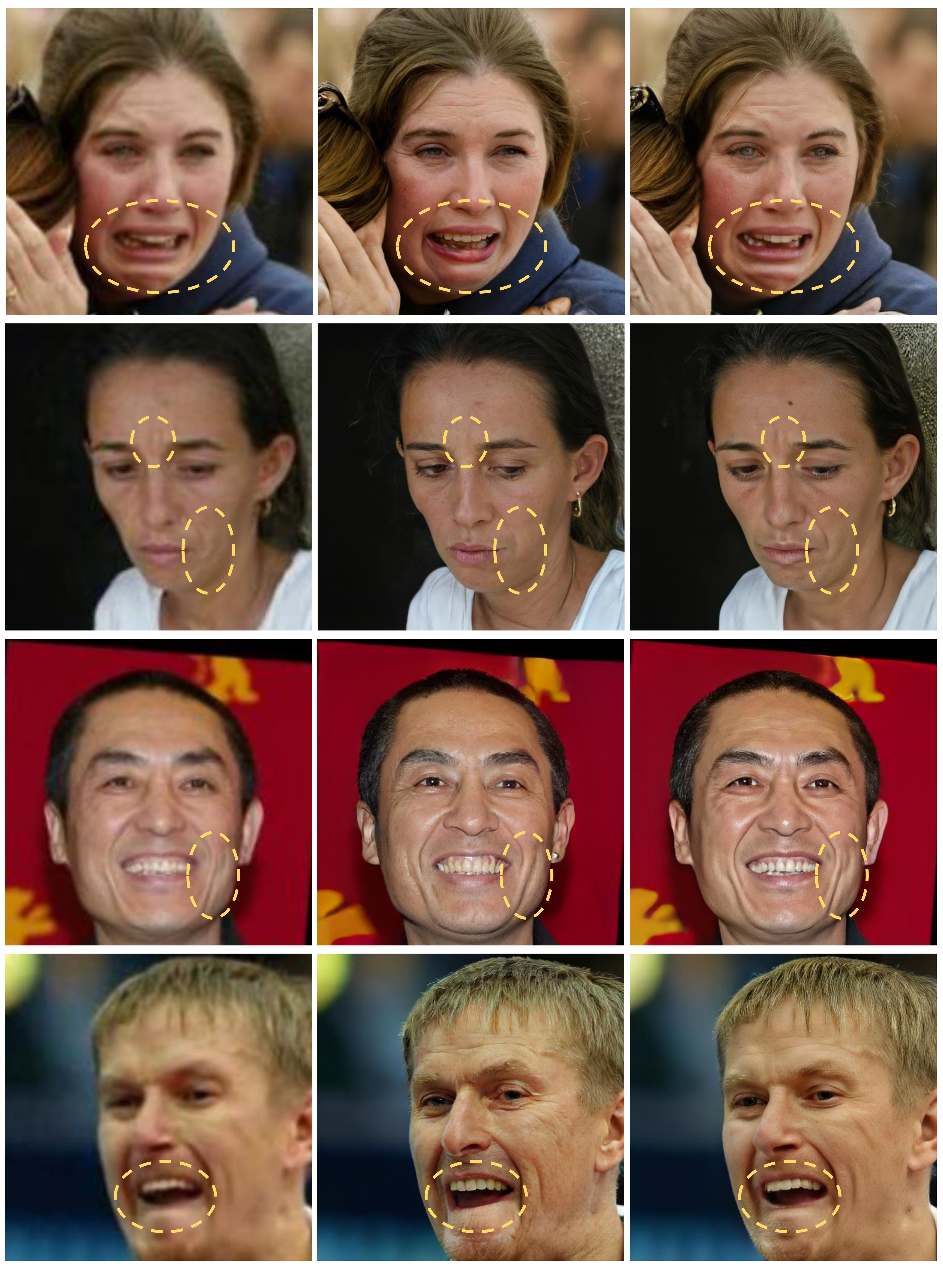}
		\put(11,0){Input}
		\put(26.7,0){SimVQFR (w/o inp feat.)}
		\put(54,0){VQFR (w/ inp feat.)}
	\end{overpic}
	\caption{Comparisons between the SimVQFR (without input features) and our VQFR (with input features). With input features of degraded faces, our VQFR could generate more faithful expressions (the first and third row), facial lines (the second and fourth row) and facial components (the fourth row) than SimVQFR.}
	\label{fig:compinpfeat}
\end{figure*}

\myPara{Importance of input features of degraded faces.}
Input features of degraded faces play an important role in preserving fidelity.
We compare the SimVQFR without input features and our VQFR with input features.
As shown in Fig.~\ref{fig:compinpfeat}, with the input features of degraded faces, VQFR could generate more faithful expressions (the first and fourth row), more faithful facial lines (the second and forth row), and facial components (the third row) than SimVQFR.
The facial lines and components can be roughly recovered from the input LQ faces but can be easily changed by the discrete quantization, thus influencing the final recovered expressions and identity.
Our VQFR incorporates the input features from degraded faces at different spatial levels and preserve better fidelity.

\begin{figure*}[!tb]
	\centering
	\begin{overpic}[width=\textwidth]{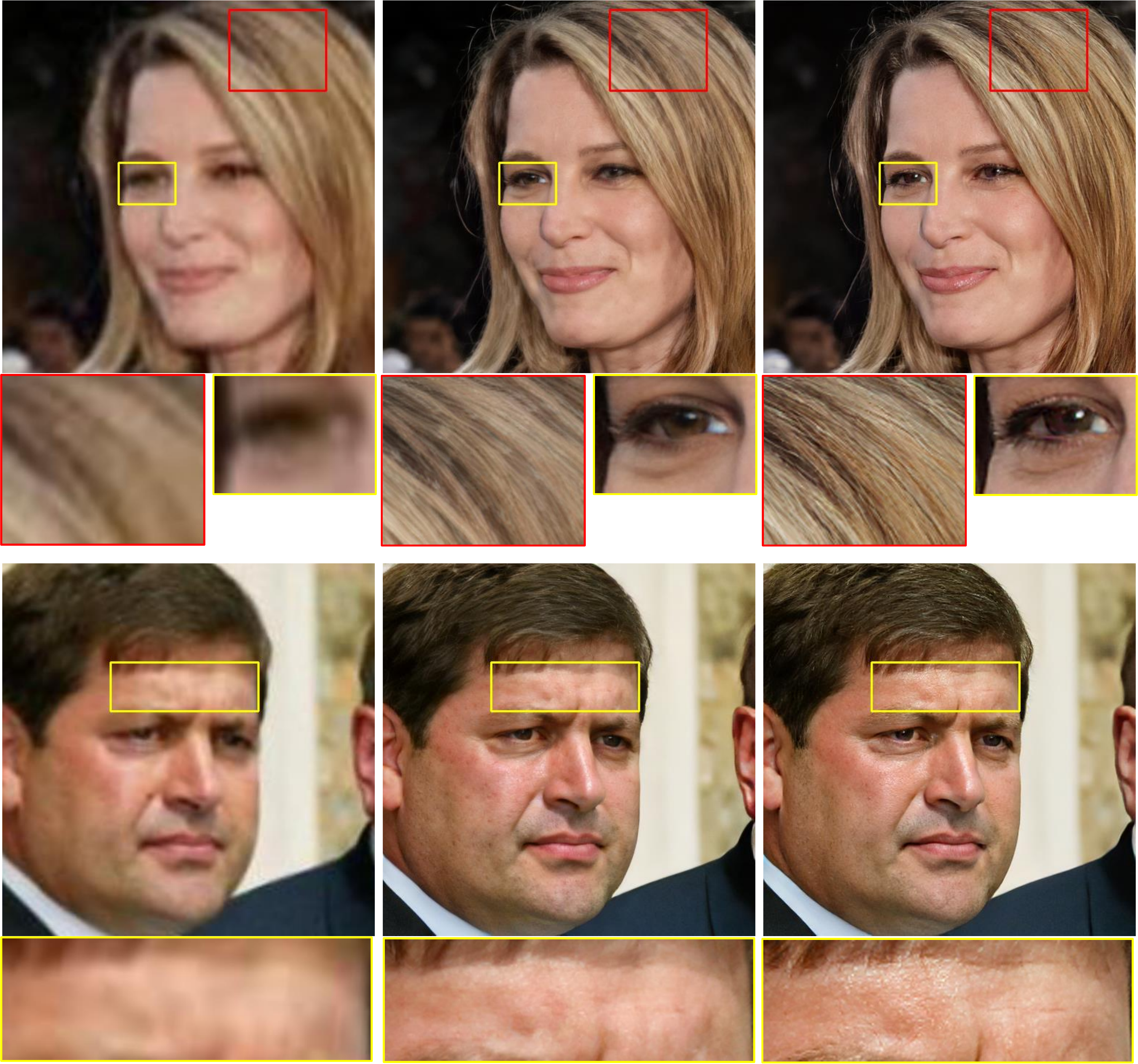}
		\put(14,-2.5){Input}
		\put(37.5,-2.5){Variant-1 (Single Branch)}
		\put(69,-2.5){Variant-2 (Parallel Decoder)}
	\end{overpic}
	\vspace{.05in}
	\caption{Comparisons between the Variant-1 (single branch) and Variant-2 (parallel decoder). With the proposed parallel decoder, high-quality facial details from the VQ codebook could be preserved. Therefore, Variant-2 could generate better facial components (the first row), more realistic hairs (the first row) and skin (the second row) than Variant-1.}
	\label{fig:focuspara}
\end{figure*}

\begin{figure*}[!tb]
	\centering
	\begin{overpic}[width=\textwidth]{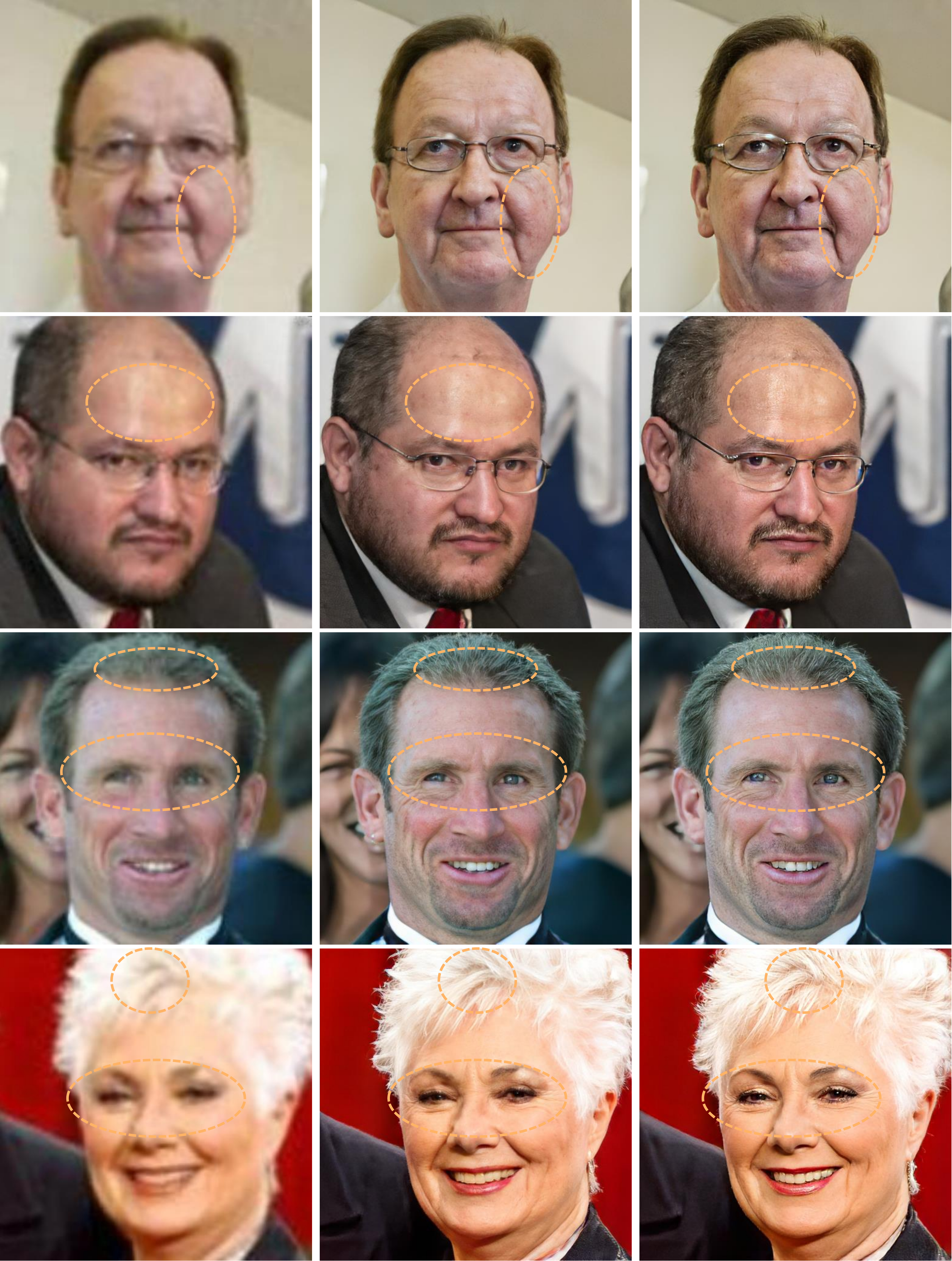}
		\put(12,-2){Input}
		\put(28,-2){Variant-1 (Single Branch)}
		\put(52,-2){Variant-2 (Parallel Decoder)}
	\end{overpic}
	\vspace{.1in}
	\caption{Comparisons between the Variant-1 (single branch) and Variant-2 (parallel decoder). With the parallel decoder, Variant-2 could generate more realistic skins (the first and second rows), eyes (the third and fourth rows) and hairs (the third and fourth rows) than Variant-1.}
	\label{fig:comppara}
\end{figure*}

\myPara{Importance of the parallel decoder.}
The parallel decoder is the key design of VQFR to preserve high-quality facial details when fusing texture features of the VQ codebook and input features from degraded faces. We compare the variant-1 (single branch) and variant-2 (parallel decoder) in \figref{fig:focuspara}. With the parallel decoder, variant-2 could generate high-quality facial components (the first row), realistic hairs (the first row) and skins (the second row).
In \figref{fig:comppara}, we provide more visual examples to show the importance of the parallel decoder design in generating realistic skins (the first and second rows) high-quality hairs and eyes (the third and fourth rows).

\begin{figure*}[!tb]
	\centering
	\begin{overpic}[width=\textwidth]{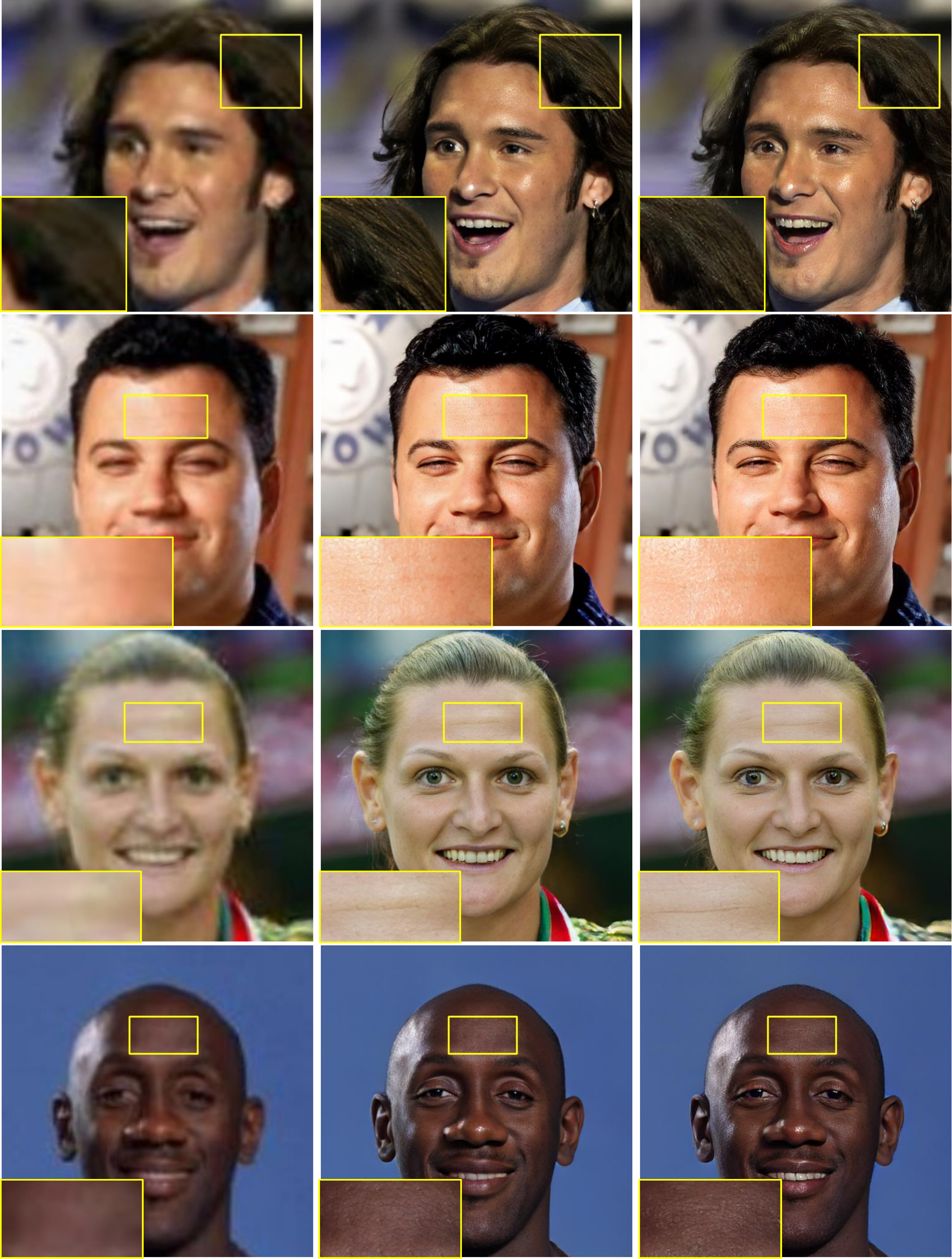}
	\put(10,-2){\large Input}
	\put(34,-2){\large Global D}
	\put(53,-2){\large Global D + Local D}
	\end{overpic}
	\vspace{.1in}
	\caption{Influence of dual discriminators. With only the global discriminator (the second column), there are regular patterns on hairs and skins. When adding the patch discriminator as a local discriminator, the regular patterns are removed (the third column).}
	\label{fig:dualD}
\end{figure*}

\myPara{Influence of dual discriminators.}
We adopt dual discriminators to remove the regular pattern when utilizing facial textures of the VQ codebook in VQFR. We adopt style-based wavelet-driven discriminator~\cite{gal2021swagan} as the global discriminator and adopt PatchGAN discriminator~\cite{isola2017image} as the local discriminator.
We show the influence of dual discriminators in \figref{fig:dualD}. When we only use the global discriminator (the second column), we can find that there are regular patterns on skin and hair. When adding the patch discriminator as the local discriminator,
the regular patterns are removed (the third column).

\clearpage
\newpage

\subsection{More Qualitative Results on Real-World Data}
\label{sec:viscompare}
We show more qualitative results on the real-world dataset, \textit{i.e.}, \textit{LFW-Test}, \textit{CelebChild} and \textit{WebPhoto}. We compare our VQFR with several state-of-the-art face restoration methods: DFDNet~\cite{li2020blind}, PSFRGAN~\cite{chen2021progressive}, PULSE~\cite{menon2020pulse} and GFPGAN~\cite{wang2021towards}.

\begin{figure*}[!tb]
	\centering
	\includegraphics[width=\linewidth]{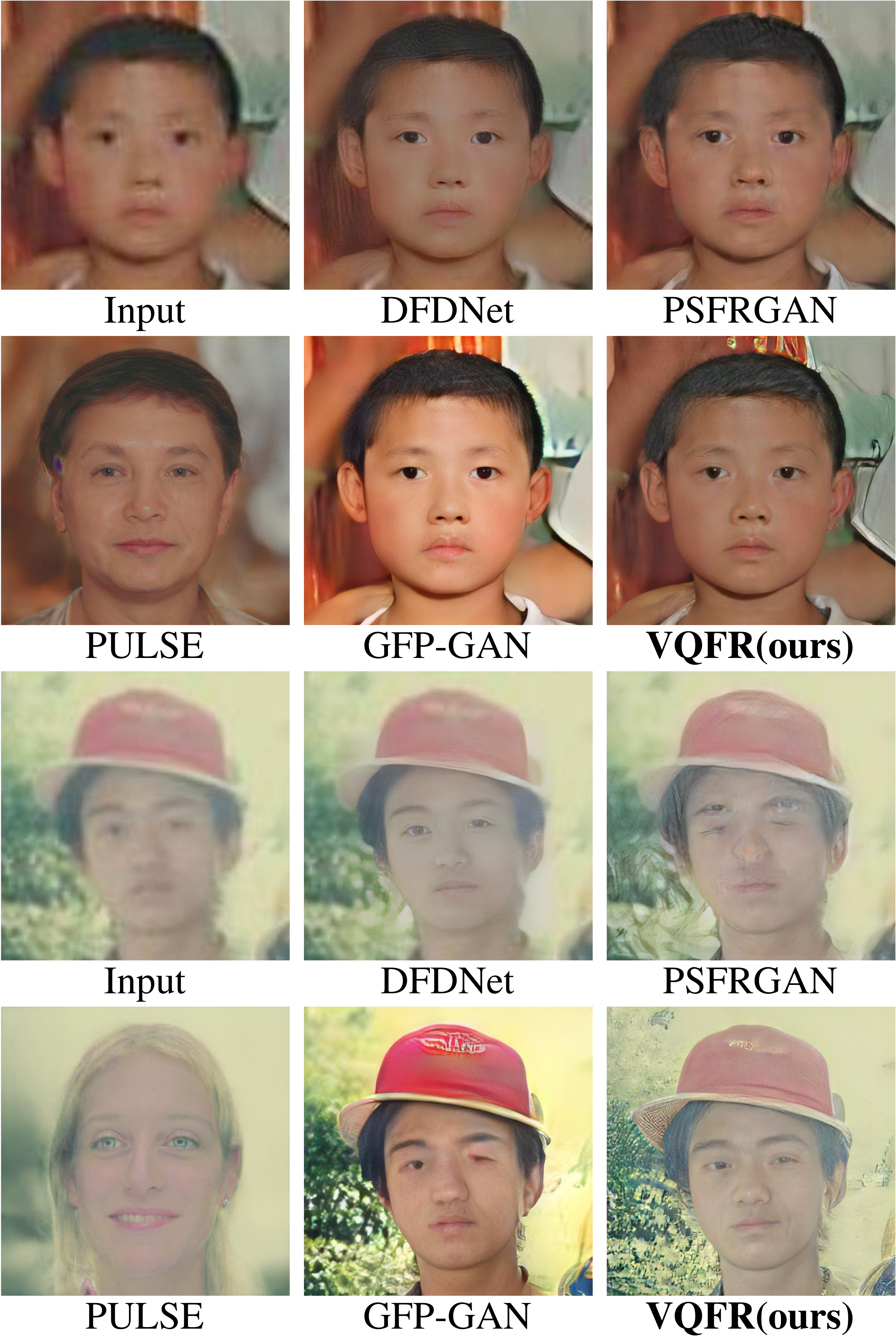}
	\caption{Qualitative comparison on the real-world \textit{WebPhoto} dataset. Our VQFR could restore more realistic facial components (eyes and ears) than previous methods. (\textbf{Zoom in for best view}).}
	\label{fig:web1}
\end{figure*}

\begin{figure*}[!tb]
	\centering
	\includegraphics[width=\linewidth]{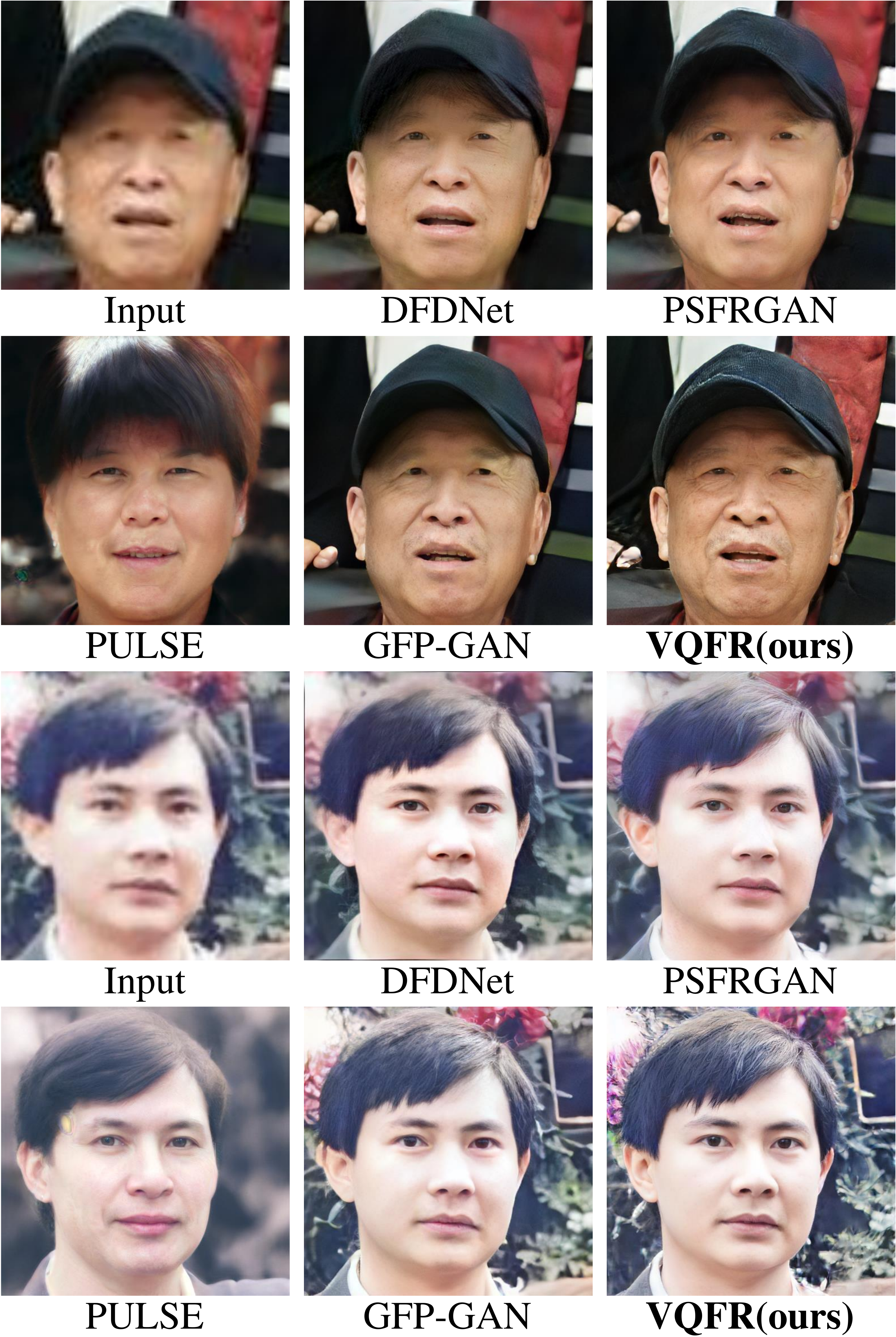}
	\caption{Qualitative comparison on the real-world \textit{WebPhoto} dataset. Our VQFR could restore more realistic facial components (eyes and ears) and more realistic skins than previous methods. (\textbf{Zoom in for best view}).}
	\label{fig:web2}
\end{figure*}

\begin{figure*}[!tb]
	\centering
	\includegraphics[width=\linewidth]{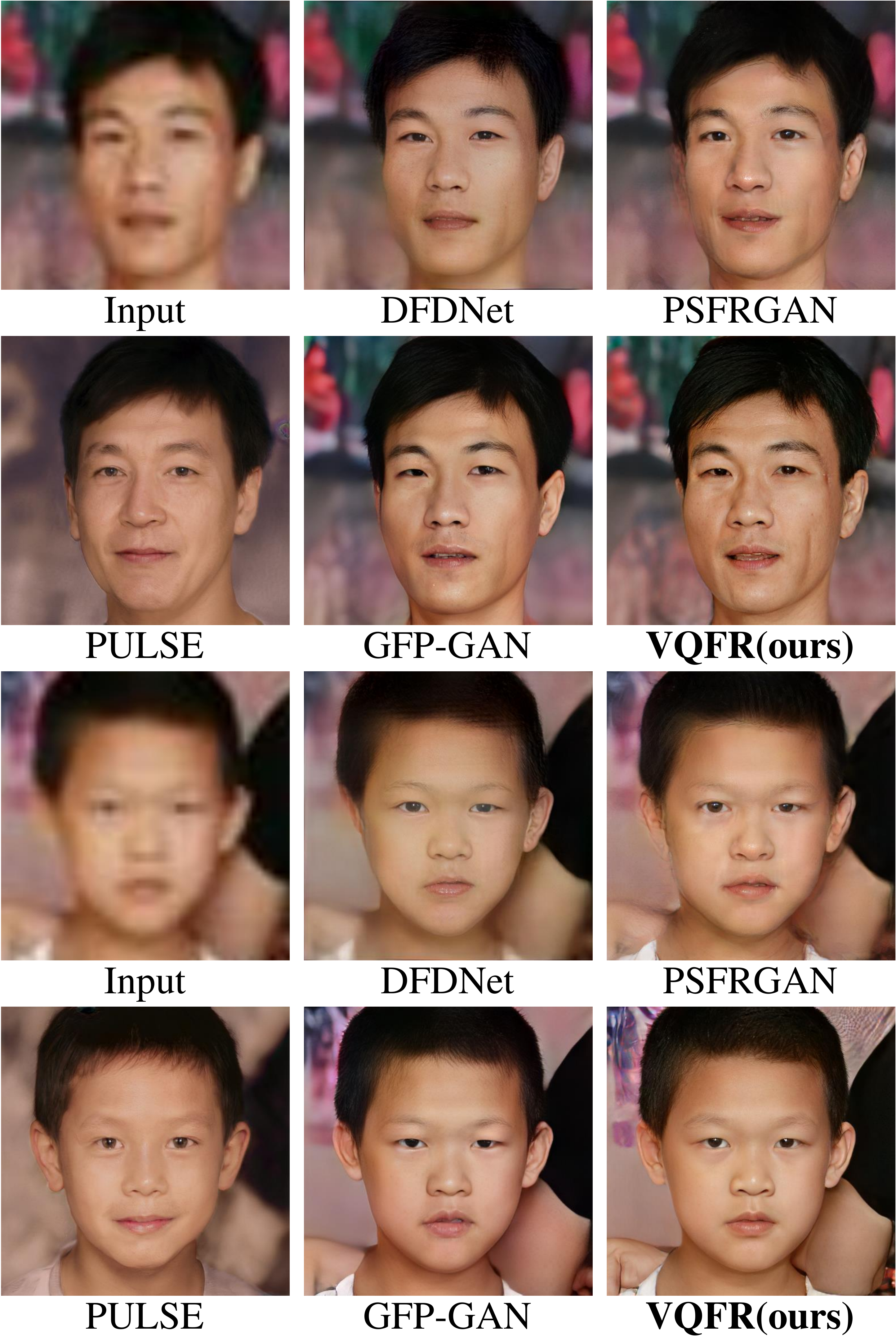}
	\caption{Qualitative comparison on the real-world \textit{WebPhoto} dataset. Our VQFR could restore more realistic facial components (eyes and ears) and more realistic skins and hairs than previous methods. (\textbf{Zoom in for best view}).}
	\label{fig:web3}
\end{figure*}

\begin{figure*}[!tb]
	\centering
	\includegraphics[width=\linewidth]{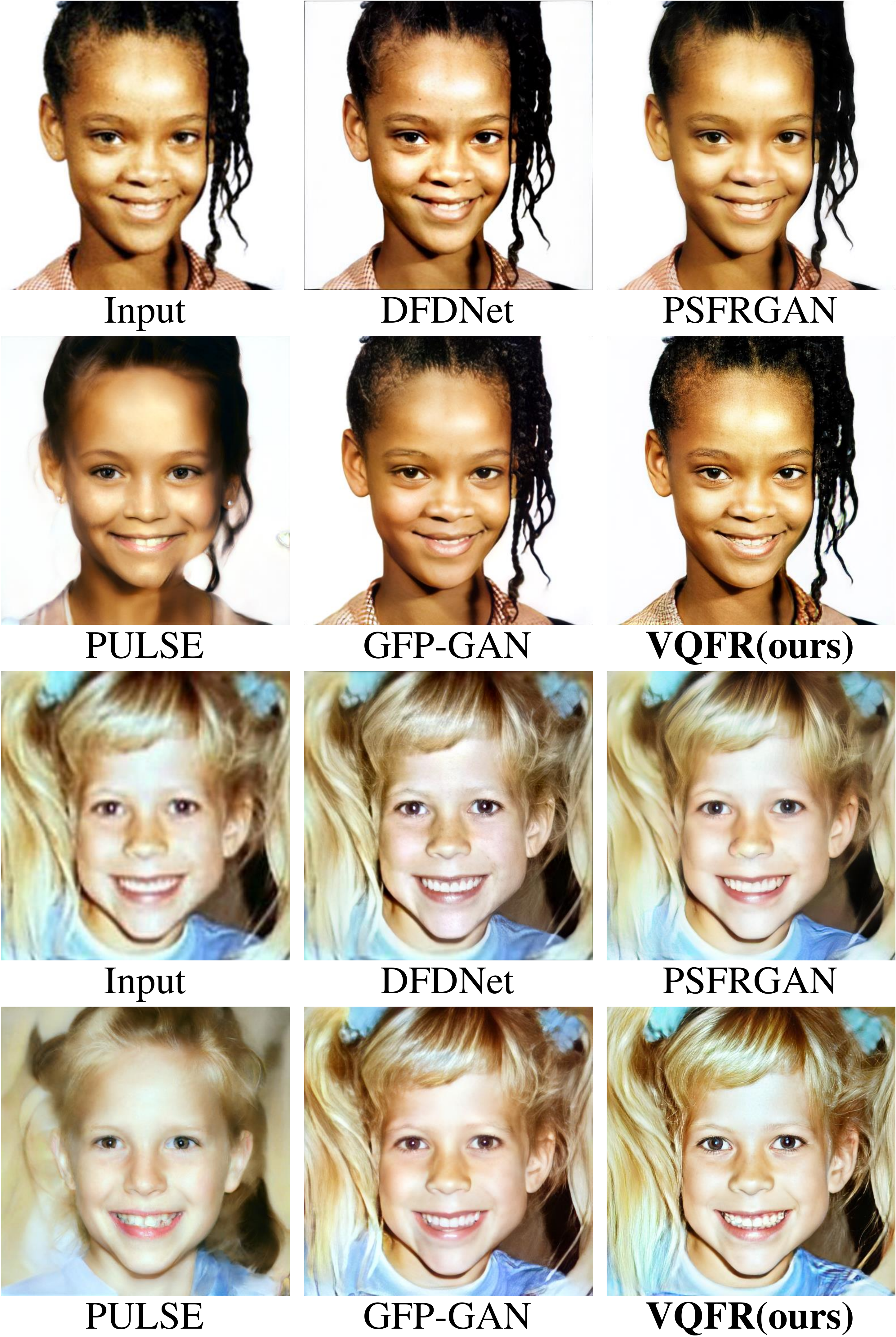}
	\caption{
	Qualitative comparison on the real-world \textit{Celeb-Child} dataset. Our VQFR could restore more realistic eyes and hairs than previous methods. (\textbf{Zoom in for best view}).}
	\label{fig:child1}
\end{figure*}

\begin{figure*}[!tb]
	\centering
	\includegraphics[width=\linewidth]{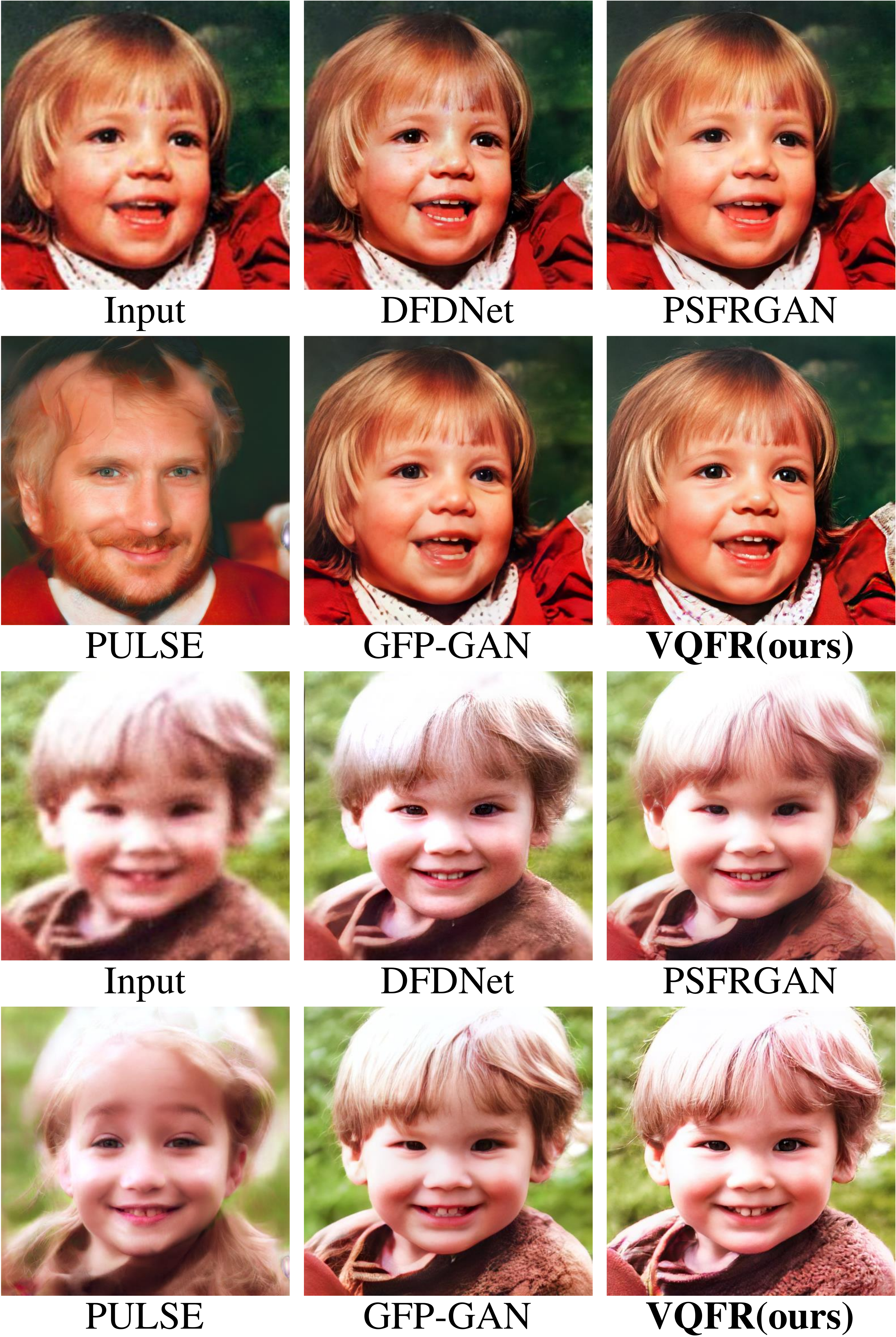}
	\caption{
	Qualitative comparison on the real-world \textit{Celeb-Child} dataset. Our VQFR could restore more realistic eyes and hairs than previous methods. (\textbf{Zoom in for best view}).}
	\label{fig:child2}
\end{figure*}

\begin{figure*}[!tb]
	\centering
    \includegraphics[width=\linewidth]{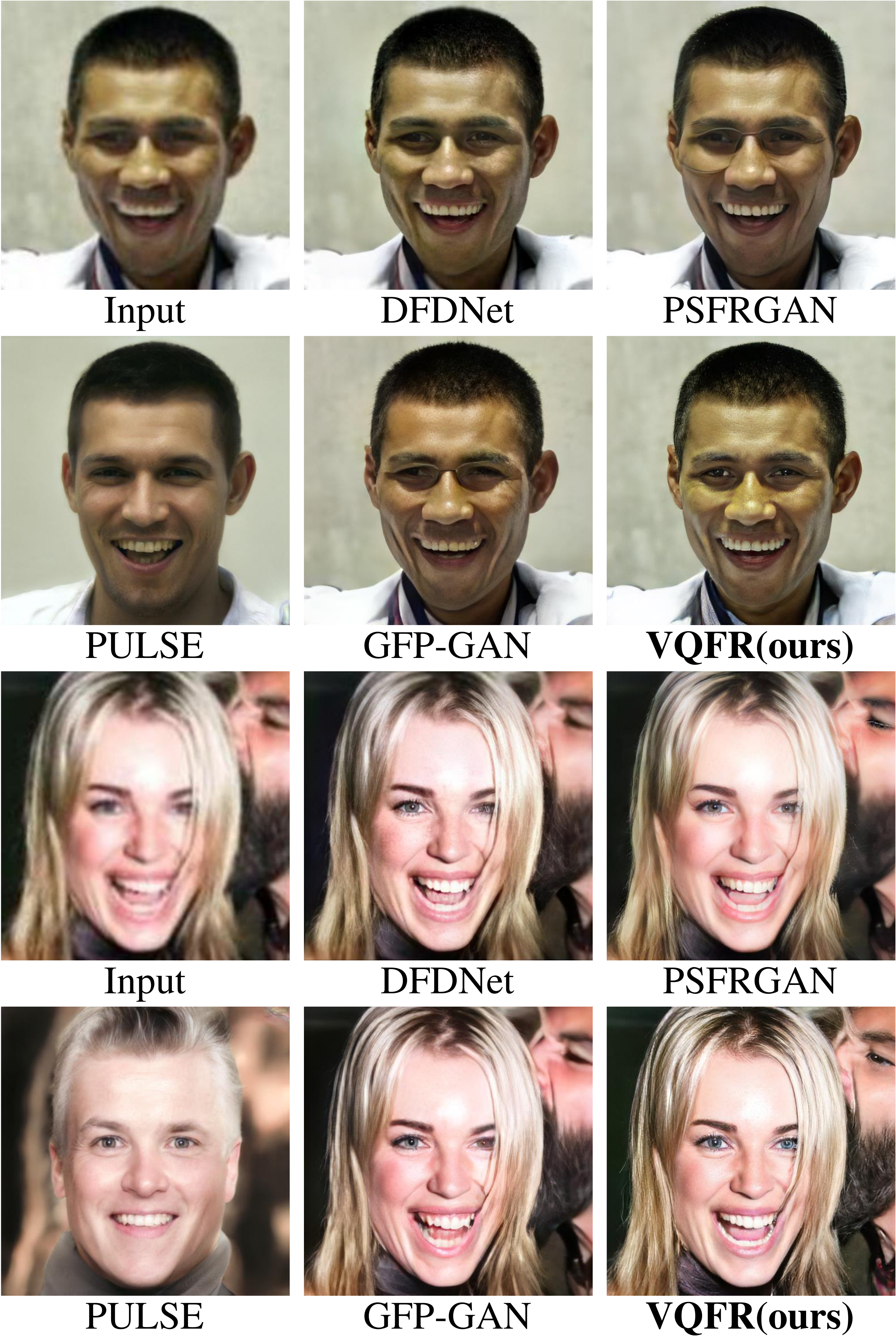}
	\caption{Qualitative comparison on the real-world \textit{LFW-Test} dataset. Our VQFR could restore high-quality facial components (eyes and hairs) and more realistic skins than previous methods. (\textbf{Zoom in for best view}).}
	\label{fig:lfw1}
\end{figure*}

\begin{figure*}[!tb]
	\centering
	\includegraphics[width=\linewidth]{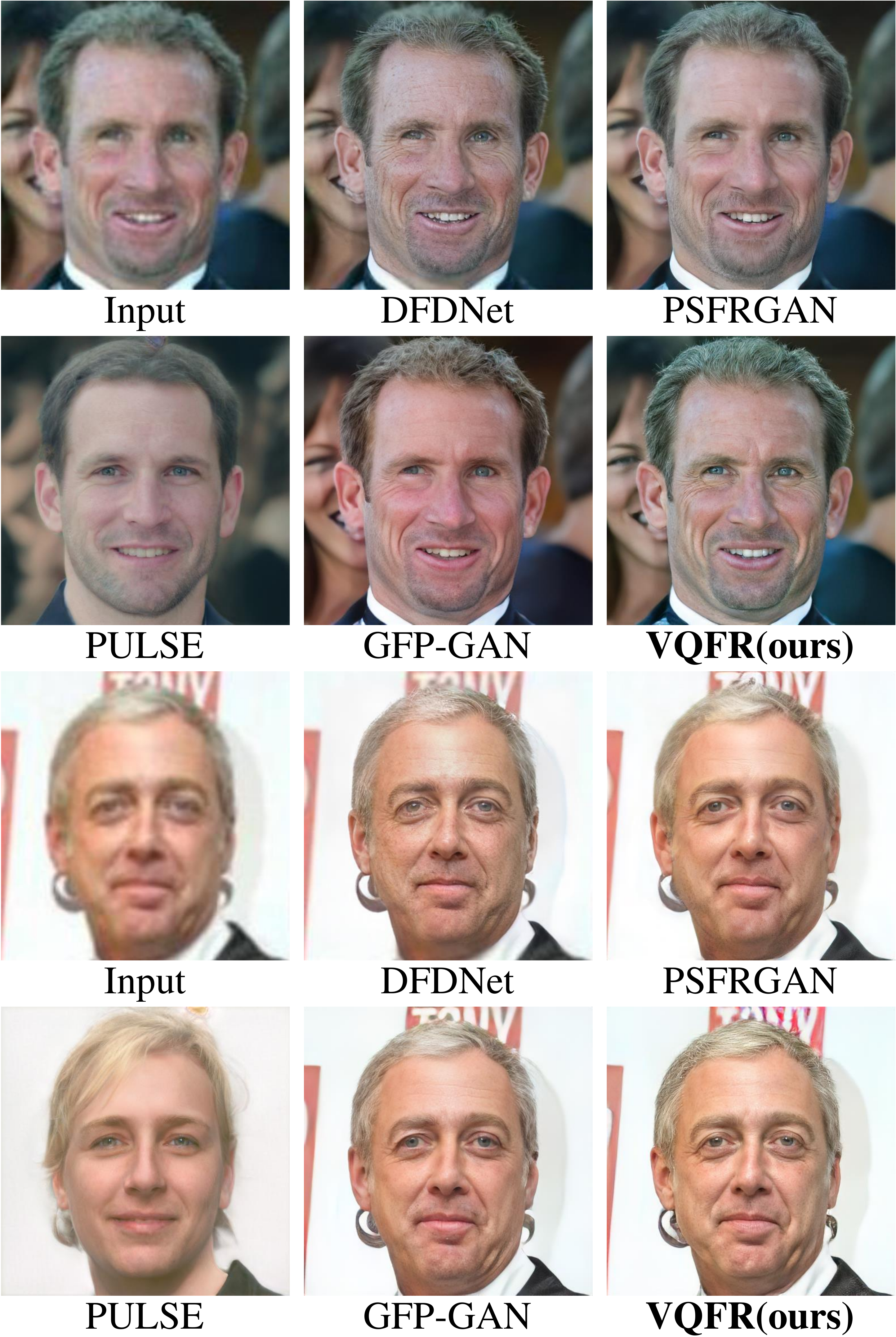}
	\caption{Qualitative comparison on the real-world \textit{LFW-Test} dataset. Our VQFR could restore high-quality facial components (eyes) and more realistic skins than previous methods. (\textbf{Zoom in for best view}).}
	\label{fig:lfw2}
\end{figure*}

\begin{figure*}[!tb]
	\centering
	\includegraphics[width=\linewidth]{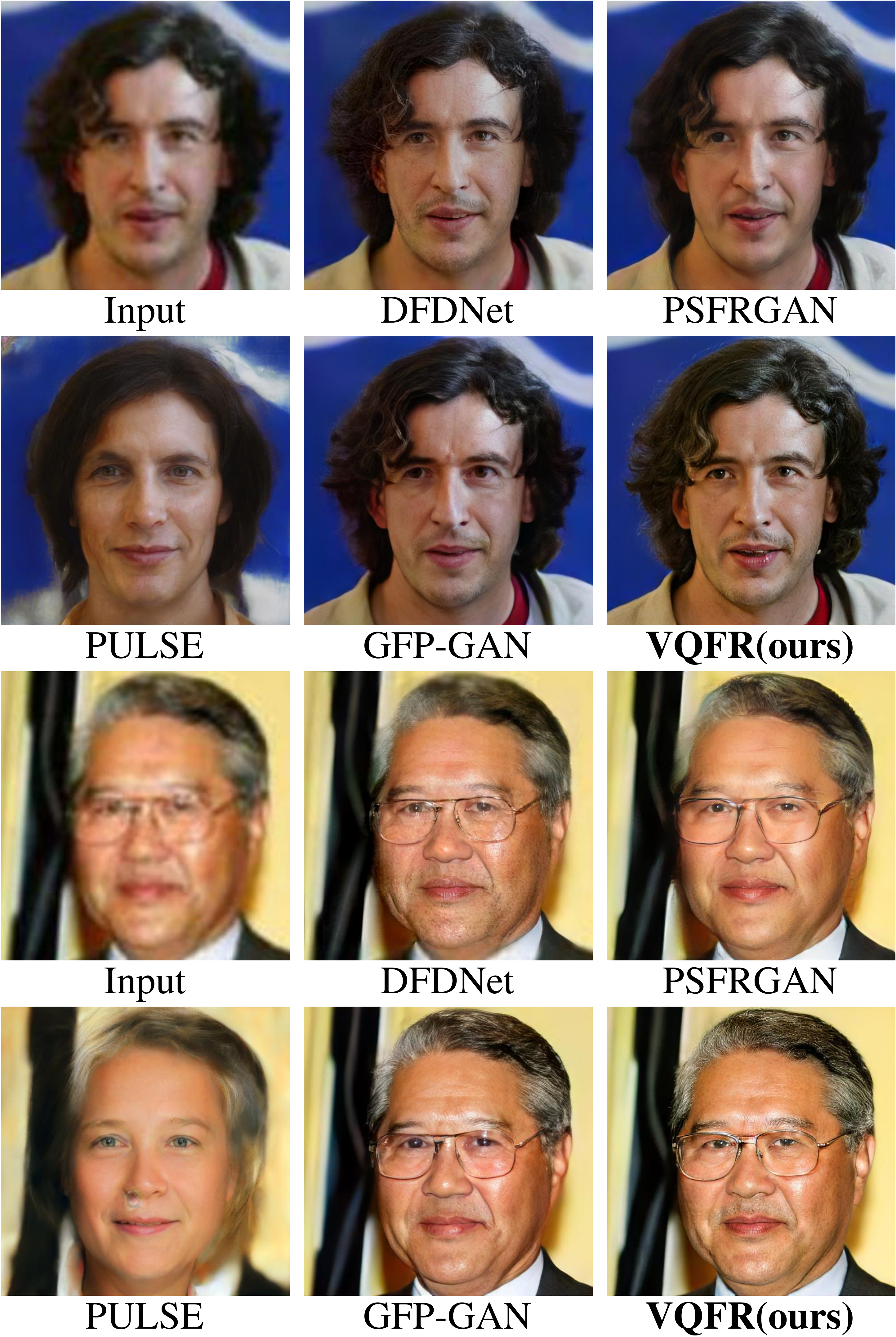}
	\caption{Qualitative comparison on the real-world \textit{LFW-Test} dataset. Our VQFR could restore high-quality facial components (eyes) and more realistic skins and hairs than previous methods. (\textbf{Zoom in for best view}).}
	\label{fig:lfw3}
\end{figure*}

The qualitative comparisons on the \textit{WebPhoto} are shown in \figref{fig:web1}, \figref{fig:web2} and \figref{fig:web3}. The qualitative comparisons on the \textit{CelebChild} are present in \figref{fig:child1} and \figref{fig:child2}. Moreover, qualitative comparisons on \textit{LFW-Test} are shown in \figref{fig:lfw1}, \figref{fig:lfw2} and \figref{fig:lfw3}.
Our VQFR produces high-quality facial components and more realistic hairs and skins than previous methods.

\end{document}